\documentclass{article}

\usepackage{microtype}
\usepackage{graphicx}
\usepackage{subcaption}
\usepackage{booktabs} 

\usepackage{hyperref}

\usepackage[preprint]{icml2026}

\usepackage{amsmath}
\usepackage{amssymb}
\usepackage{mathtools}
\usepackage{amsthm}
\usepackage{multirow}
\usepackage[table]{xcolor}
\definecolor{skyblue}{rgb}{0.26, 0.52, 0.73}
\definecolor{lightcoral}{rgb}{0.8, 0.3, 0.3}

\DeclareRobustCommand{\github}{%
  \begingroup
  \raisebox{-0.2em}{%
  \includegraphics[height=1em]{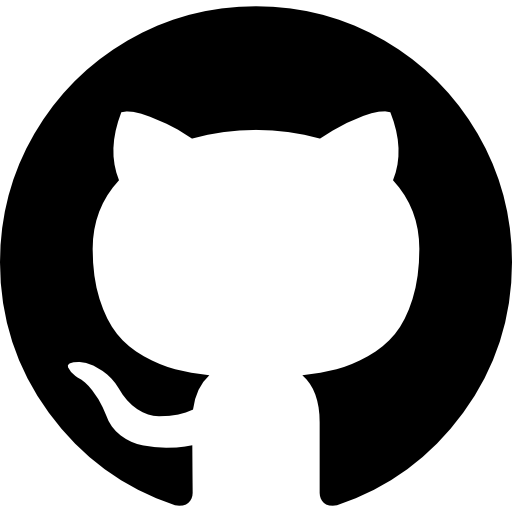}%
  }%
  \kern 0.2em%
  \endgroup
}

\usepackage[capitalize,noabbrev]{cleveref}

\theoremstyle{plain}

\theoremstyle{definition}

\theoremstyle{remark}

\usepackage[textsize=tiny]{todonotes}

\icmltitlerunning{Machine Text Detectors are Membership Inference Attacks}

\begin{document}

\twocolumn[
  \icmltitle{Machine Text Detectors are Membership Inference Attacks}



  \icmlsetsymbol{equal}{*}

  \begin{icmlauthorlist}
    \icmlauthor{Ryuto Koike}{equal,st,penn}
    \icmlauthor{Liam Dugan}{equal,penn}
    \icmlauthor{Masahiro Kaneko}{mbzuai}
    \icmlauthor{Chris Callison-Burch}{penn}
    \icmlauthor{Naoaki Okazaki}{st}
  \end{icmlauthorlist}
  \icmlaffiliation{st}{Institute of Science Tokyo}
  \icmlaffiliation{penn}{University of Pennsylvania}
  \icmlaffiliation{mbzuai}{Mohamed bin Zayed University of Artificial Intelligence}
  \icmlcorrespondingauthor{Ryuto Koike}{ryuto.koike@nlp.c.titech.ac.jp}
  \icmlcorrespondingauthor{Liam Dugan}{ldugan@seas.upenn.edu}

  \icmlkeywords{Machine Learning, ICML}

  \vskip 0.3in
]



\printAffiliationsAndNotice{*Equal contribution; Ryuto Koike did his work while visiting the University of Pennsylvania.}  

\begin{abstract}
Although membership inference attacks (MIAs) and machine-generated text detection target different goals, their methods often exploit similar signals based on a language model’s probability distribution, and the two tasks have been studied independently. This can result in conclusions that overlook stronger methods and valuable insights from the other task.
In this work, we theoretically and empirically demonstrate the \textit{transferability}, i.e., how well a method originally developed for one task performs on the other, between MIAs and machine text detection.
We prove that the metric achieving asymptotically optimal performance is identical for both tasks. We unify existing methods under this optimal metric and hypothesize that the accuracy with which a method approximates this metric is directly correlated with its transferability.
Our large-scale empirical experiments demonstrate very strong rank correlation ($\rho \approx 0.7$) in cross-task performance.
Notably, we also find that a machine text detector achieves the strongest performance among evaluated methods on both tasks, demonstrating the practical impact of transferability.
To facilitate cross-task development and fair evaluation, we introduce \textsc{Mint}, a unified evaluation suite for MIAs and machine-generated text detection, implementing 15 recent methods from both tasks. \github\url{https://github.com/ryuryukke/mint}
\end{abstract}


\begin{figure*}[t]
 \small
\centering\includegraphics[width=0.9\textwidth]{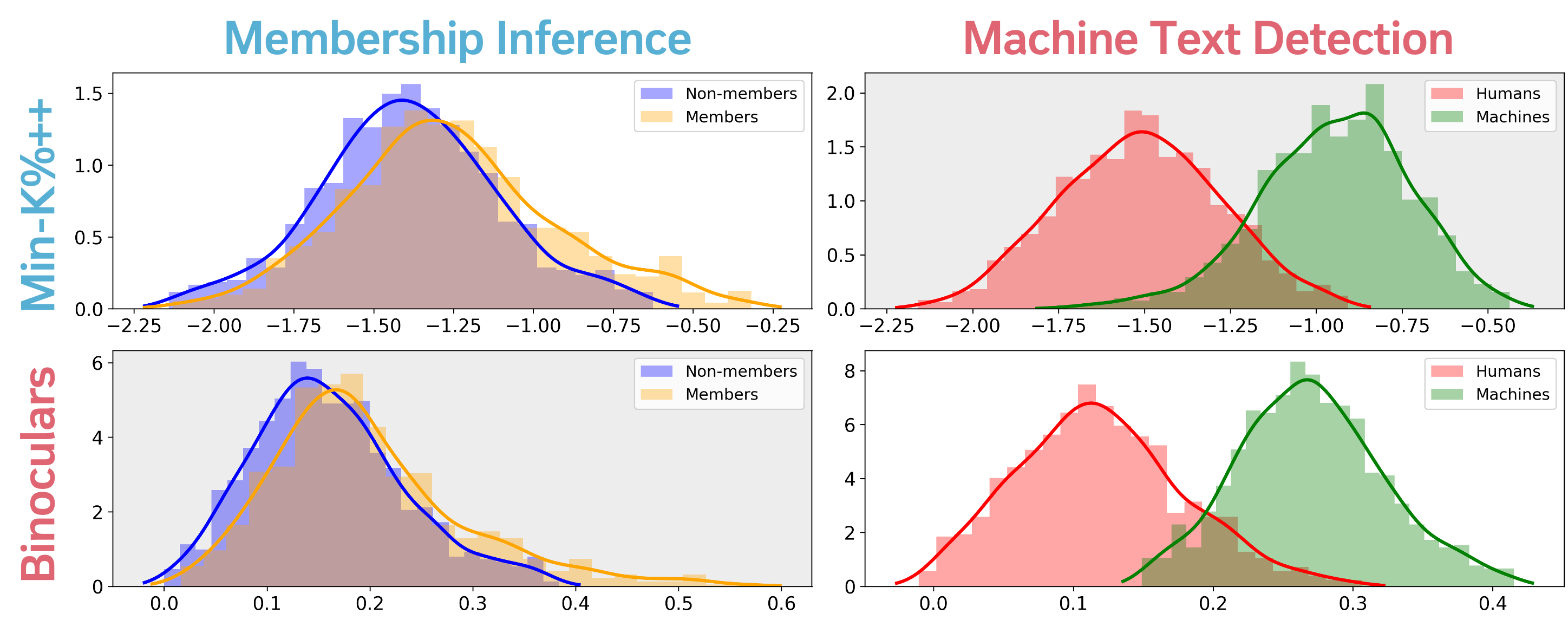}
\caption{Predicted score distributions of \textit{Min-K\%++} (state-of-the-art MIA) and \textit{Binoculars} (state-of-the-art machine text detector) on both tasks. Shaded areas indicate the cross-task setting. Although both methods were developed for two separate tasks, the distributions across populations induced by both methods are \textbf{strikingly similar}, suggesting their transferability.}
  \label{cross_task_distribution_plot}
\end{figure*}

\section{Introduction}
Large language models (LLMs) have demonstrated human-level generative and understanding capabilities, impacting fields such as creative writing \citep{10.1145/3613904.3642731}, news reporting \citep{AI_news_reporting}, and even scientific discovery \citep{lu2024aiscientist}. Despite many positive societal implications, their negative consequences have increasingly been reported. For instance, LLMs may leak personal \citep{lukas2023analyzing} or copyrighted information \citep{NEURIPS2024_faed4276} due to their memorization of training data \citep{morris2025languagemodelsmemorize, carlini-etal-2021}. In addition to privacy issues, LLMs also raise challenges to authorship authenticity, as they can be exploited for mass-producing propaganda \citep{10.1093/pnasnexus/pgae034} or cheating on student assignments \citep{chatgpt_cheating}.

Many recent works aim to mitigate such negative implications of LLMs. One major direction is membership inference attacks (MIAs), which aim to classify whether a given text sample is a member of a language model's training data \citep{carlini-etal-2022,mattern-etal-2023-membership,min_k}. This helps identify potential leaks of personal information and copyright infringement. To address the challenges of authorship authenticity, another major line of work is machine-generated text detection, which distinguishes between human-written and machine-generated texts \citep{ippolito-etal-2020-automatic,pmlr-v202-mitchell23a,binoculars,yang-etal-2024-survey}. This safeguards against misuse of LLMs (e.g., misinformation and academic misconduct) by flagging suspicious texts.

While these two tasks, MIAs and machine text detection, target different goals, their methods often leverage similar signals based on a language model's probability distribution. In MIAs, documents that are members of a model's training dataset tend to show higher likelihoods under the model than non-members. Similarly, in machine text detection, machine-generated texts typically exhibit higher likelihoods under the target model than human-written texts. Thus, both tasks use text likelihood or entropy as standard baselines. Indeed, as illustrated in Figure \ref{cross_task_distribution_plot}, our pilot study shows strikingly similar score distributions of the state-of-the-art MIA and detector across both tasks. Despite this shared property, the two tasks have been studied independently. This can result in biased evaluations that overlook stronger methods developed for one task and consequently lead to conclusions that miss valuable insights from the other.


Motivated by this gap, we theoretically and empirically study the \textit{transferability}, i.e., how well a method originally developed for one task performs on the other, between MIAs and machine text detection.
For our theoretical contribution, we prove that both tasks share the same optimal metric for achieving asymptotically highest performance: \textit{the likelihood ratio test between the target model distribution and the true population distribution} (\S\ref{unified_framework}). We unify a large proportion of the existing methods from both tasks as approximations of this optimal metric and hypothesize that the transferability of a method is correlated with how well it approximates this metric.

To empirically quantify the transferability, we conduct large-scale experiments, spanning 7 state-of-the-art MIA methods and 5 state-of-the-art machine text detectors across 13 domains and 10 generators (\S\ref{quantify_transferability}). For each method, we evaluate its performance on both tasks, MIAs and machine text detection, and then compute the rank correlation between their performance rankings on the two tasks. We find that many methods effective in MIAs remain effective in machine text detection, and vice versa, with a strong rank correlation ($\rho \approx 0.7$) in their cross-task performance (\S\ref{results}).
Furthermore, we notably find that \textit{Binoculars} \citep{binoculars}, originally designed for machine text detection, achieves state-of-the-art performance on MIAs as well, demonstrating the practical impact of the transferability. 

\paragraph{Contributions.}
Our contributions are as follows: (1) We are the first to provide proof that MIAs and machine text detection share the same asymptotically optimal metric. This identifies what determines performance in both tasks, providing practical guidance for developing stronger methods. (2) We show that many popular methods can be viewed as approximations to this shared optimal metric, which explains when and why transferability arises and how its strength depends on the degree of this approximation. (3) We present the first comprehensive empirical study of transferability between the two tasks, showing strong rank correlation in cross-task performance. We further find that a machine text detector can outperform state-of-the-art MIA methods even on the MIA benchmark, highlighting the need for fair and unified evaluation.

\section{Theoretical Transferability between MIAs and Machine Text Detection}

\subsection{Task Formulation}
Let $\mathcal{X}$ denote the set of all token sequences. 
Let $\mathcal{Q}$ be an oracle that models the ``true'' probability distribution $P_{\mathcal{Q}}$ of human-written text over $\mathcal{X}$. 
Let $\mathcal{M}$ be a language model that induces a distribution $P_{\mathcal{M}}$ over $\mathcal{X}$, and let $\mathcal{D}_{\text{train}} \subset \mathcal{X}$ denote the unknown training set of $\mathcal{M}$.

\paragraph{Machine-Generated Text Detection.} The objective is to determine whether a given text $x \in \mathcal{X}$ is written by humans or generated by the model $\mathcal{M}$. This can be formalized as the following hypothesis test
\[
\begin{aligned}
&H_0: x \sim P_{\mathcal{Q}},\;H_1: x \sim P_{\mathcal{M}}.\\
\end{aligned}
\]
\paragraph{Membership Inference.} 
The objective is to determine whether or not the model $\mathcal{M}$ was trained on a given text $x \in \mathcal{X}$.
Let $\mathcal{T}$ denote a training algorithm and let $\mathcal{D}' \sim P_{\mathcal{Q}}^{N}$ be a random training dataset sampled from $P_{\mathcal{Q}}$. This can be formalized as the following hypothesis test, which differs only in whether $x$ is included during training,
\[
\begin{aligned}
&H_0: \mathcal{M} \leftarrow \mathcal{T}(\mathcal{D}'),\; H_1: \mathcal{M} \leftarrow \mathcal{T}(\mathcal{D}'\cup\{x\}),\\
\end{aligned}
\]

The goal of both tasks is to design a test statistic $f(x;\mathcal{M})$ that rejects the respective null hypothesis $H_0$ with maximal statistical power.

\subsection{Unifying MIAs and Machine Text Detection as Likelihood Ratio Tests}
\label{unified_framework}
\textbf{Theorem 2.2 (Unified Optimality).} 
Let $\mathcal{X}$ denote the set of all token sequences. Let $\mathcal{M}$ be a language model trained to maximize the likelihood of a training set $\mathcal{D}_{\text{train}} \subset \mathcal{X}$, and let $P_{\mathcal{M}}$ denote the probability distribution induced by $\mathcal{M}$. Let $P_{\mathcal{Q}}$ denote a probability distribution over the subset of $\mathcal{X}$ that models the ``true'' human distribution of text. 
Then the test statistic
\[
\Lambda(x) = \frac{P_{\mathcal{M}}(x)}{P_{\mathcal{Q}}(x)}
\]
achieves optimal accuracy at a given false positive rate (Type I error) \textit{for both machine-generated text detection and membership inference} under standard asymptotic regularity conditions, with corresponding maximum advantage (improvement over random guessing) bounded by
\[
\operatorname{adv} \leq \sqrt{\frac{D_{\mathrm{KL}}(P_{\mathcal{Q}} \| P_{\mathcal{M}})}{8}}.
\]

\textbf{Proof.} We prove this by showing that $\Lambda(x)$ coincides with the likelihood ratio test for both tasks.

\textit{Step 1: Machine-generated text detection.}  
Consider testing
\[
H_0: x \sim P_{Q}, \;\; H_1: x \sim P_{M}.
\]
The likelihood ratio test for this hypothesis is
\[
\Lambda_{\text{MGT}}(x) = \frac{P_{\mathcal{M}}(x)}{P_{\mathcal{Q}}(x)}
\]
which exactly coincides with the proposed statistic $\Lambda(x)$. By the Neyman-Pearson lemma \citep{neyman1933problem}, this test statistic is most powerful at a given Type I error.

\textit{Step 2: Membership inference.}
Consider testing
\[
\begin{aligned}
&H_0: \mathcal{M} \leftarrow \mathcal{T}(\mathcal{D}'), \;\;
H_1: \mathcal{M} \leftarrow \mathcal{T}(\mathcal{D}' \cup \{x\}).
\end{aligned}
\]
Following \citet{carlini-etal-2022}, membership inference can be formulated as a likelihood
ratio test between the distributions over trained models with and without the target
data point $x$,
\[
\Lambda_{\text{MI}}(x)
=
\frac{P(\mathcal{M} \mid H_1)}{P(\mathcal{M} \mid H_0)},
\]
where $P(\mathcal{M} \mid H)$ denotes the likelihood evaluated at the observed trained
model under the corresponding training hypothesis.
The exact likelihood over models is intractable.
Under our assumptions, however, the likelihood ratio depends on the trained model only
through the scalar statistic
\[
s(x;\mathcal M) = P_{\mathcal M}(x),
\]
which is sufficient for distinguishing $H_0$ and $H_1$
(see Appendix~\ref{app:detailed_proof} for a detailed proof of this).
Accordingly, the likelihood ratio can be rewritten as
\[
\Lambda_{\text{MI}}(x)
=
\frac{P\!\left(s(x;\mathcal M)\mid H_1\right)}
     {P\!\left(s(x;\mathcal M)\mid H_0\right)}.
\]
Under the asymptotic assumption that $\mathcal M$ has infinite capacity and perfectly
maximizes the likelihood of the training set, $P_{\mathcal M}(x)$ equals the empirical
frequency of $x$ in the training set $\mathcal D_{\text{train}}$.
Let $K = N \cdot s(x;\mathcal M)$ denote the number of occurrences of $x$ in the training set, where
$N = |\mathcal D_{\text{train}}|$.
Then the likelihood ratio further reduces to
\[
\Lambda_{\text{MI}}(x)
=
\frac{P(K \mid H_1)}{P(K \mid H_0)}.
\]
Let $q_x$ be shorthand for $P_{\mathcal Q}(x)$, we have
\[
\begin{aligned}
&K \mid H_0 \sim \mathrm{Binomial}(N, q_x),\\
&K \mid H_1 \sim 1 + \mathrm{Binomial}(N-1, q_x).
\end{aligned}
\]
For any $k \in \{1,\ldots,N\}$,
\[
\frac{\Pr[K=k \mid H_1]}{\Pr[K=k \mid H_0]}
=
\frac{\binom{N-1}{k-1} q_x^{k-1}(1-q_x)^{N-k}}
     {\binom{N}{k} q_x^{k}(1-q_x)^{N-k}}
=
\frac{k}{N q_x}.
\]
Substituting $k = N \cdot p_x$ with $p_x = P_{\mathcal M}(x)$ yields
\[
\Lambda_{\text{MI}}(x) = \frac{p_x}{q_x}.
\]
Therefore, the likelihood ratio test for membership inference reduces to
\[
\Lambda_{\text{MI}}(x) = \frac{P_{\mathcal M}(x)}{P_{\mathcal Q}(x)}.
\]


\textit{Step 3: Advantage bound.}  
For a binary hypothesis test with statistic $\Lambda(x)$, the Bayes error is
\[
\varepsilon^* = \frac{1 - \mathrm{TV}(P_{\mathcal{M}}, P_{\mathcal{Q}})}{2},
\]
where $\mathrm{TV}(\cdot, \cdot)$ denotes total variation distance. Applying Pinsker's inequality,
\[
\mathrm{TV}(P_{\mathcal{M}}, P_{\mathcal{Q}}) \le \sqrt{\frac{1}{2} D_{\mathrm{KL}}(P_{\mathcal{Q}} \| P_{\mathcal{M}})},
\]
yields the stated bound on the error and the corresponding maximum advantage. \hfill$\square$

\begin{table*}[t]
\centering
\caption{Unified formulations of \textcolor{skyblue}{membership inference} and \textcolor{lightcoral}{machine text detection} methods. $P_{\mathcal{M}}(x)$ denotes text likelihood, $\mathcal{R}(x;\mathcal{M})$ denotes average log rank, $\phi(x)$ denotes an arbitrary perturbation function, and $\Phi(x) = x / \sigma_x$ denotes division by the standard deviation. See Appendix \ref{app:details_of_methods} for more details on how we derive each formula.}
\small
\setlength{\tabcolsep}{10pt}
\renewcommand{\arraystretch}{1.2}
\begin{tabular}{l|l|l}
\toprule
\textbf{Method} & \textbf{Task} & \textbf{Equational form} \\
\midrule
\textit{Reference} \cite{carlini-etal-2021} & \textcolor{skyblue}{MIA} &
$-\log (P_{\mathcal{M}}(x) / P_{\mathcal{M}_\text{ref}}(x))$ \\
\textit{Zlib} \cite{carlini-etal-2021} & \textcolor{skyblue}{MIA} &
$-\log P_{\mathcal{M}}(x) / \text{Zlib}(x)$ \\
\textit{DetectLLM} \cite{su-etal-2023-detectllm} & \textcolor{lightcoral}{Detection} &
$\mathbb{E}_{\tilde{x}\sim \phi(x)}[\log R_{\mathcal{M}}(\tilde{x})] / (\log R_{\mathcal{M}}(x))$ \\
\textit{ReCall} \cite{xie-etal-2024-recall} & \textcolor{skyblue}{MIA} &
$\mathbb{E}_{\tilde{x}\sim \phi(x)}[\log P_{\mathcal{M}}(\tilde{x})] / (\log P_{\mathcal{M}}(x))$ \\
\textit{DC-PDD} \cite{dc_pdd} & \textcolor{skyblue}{MIA} &
$\mathbb{E}_{\tilde{x}\sim\mathcal{M}}[-\log P_{\mathcal{M}_\text{ref}}(\tilde{x})]$ \\
\textit{Binoculars} \cite{binoculars} & \textcolor{lightcoral}{Detection} &
$(\log P_{\mathcal{M}}(x)) / \mathbb{E}_{\tilde{x}\sim\mathcal{M}}[\log P_{\mathcal{M}_\text{ref}}(\tilde{x})]$ \\
\textit{DetectGPT} \cite{pmlr-v202-mitchell23a} & \textcolor{lightcoral}{Detection} &
$\mathbb{E}_{\tilde{x}\sim\phi(x)}[\log (P_{\mathcal{M}}(x) / P_{\mathcal{M}}(\tilde{x}))]$ \\
\textit{Neighborhood} \cite{mattern-etal-2023-membership} & \textcolor{skyblue}{MIA} &
$\mathbb{E}_{\tilde{x}\sim\phi(x)}[\log (P_{\mathcal{M}}(x) / P_{\mathcal{M}}(\tilde{x}))]$ \\
\textit{Fast-DetectGPT} \cite{bao2023fast} & \textcolor{lightcoral}{Detection} &
$\Phi(\mathbb{E}_{\tilde{x}\sim\phi(x)}[\log (P_{\mathcal{M}}(x) / P_{\mathcal{M}}(\tilde{x}))])$ \\
\textit{Min-$K\%$} \cite{min_k} & \textcolor{skyblue}{MIA} &
$\frac{1}{k}\sum_{i\in \text{min-}k\%} (-\log P_{\mathcal{M}}(x_i))$ \\
\textit{Min-$K\%$++} \cite{min_k_plus_plus} & \textcolor{skyblue}{MIA} &
$\frac{1}{k}\sum_{i\in \text{min-}k\%}
 \Phi(-\log P_{\mathcal{M}}(x_i) + \mathbb{E}_{\tilde{x}_i\sim\mathcal{M}}[\log P_{\mathcal{M}}(\tilde{x}_i)])$ \\
\textit{Lastde} \cite{lastde} & \textcolor{lightcoral}{Detection} &
$(-\log P_{\mathcal{M}}(x)) / \mathrm{StdDev}(\{\mathrm{DE}(x,\tau)\}_{\tau=1}^{\tau'})$ \\
\textit{Lastde++} \cite{lastde} & \textcolor{lightcoral}{Detection} &
$\Phi(\mathrm{Lastde}(x) - \mathbb{E}_{\tilde{x}\sim \phi(x)}[\mathrm{Lastde}(\tilde{x})])$ \\
\bottomrule
\end{tabular}
\label{tab:metric_equations}
\end{table*}

\paragraph{Remark.}
When $\mathcal{M}$ has sufficient capacity and is trained through asymptotically many epochs on the training set, the statistic $\Lambda(x)$ on membership inference will approach the optimal bound.
In practice, however, language models are typically trained with only a single pass over the data, and other creative approximations of $P(x\in\mathcal{D}_\text{train})$ may outperform $P_{\mathcal{M}}(x)$. Our results do not discourage further work on these approximations, but rather provide a theoretical framework for understanding the fundamental connection between machine text detection and membership inference.

\paragraph{Discussion.} A key implication of this result is that any method that effectively approximates $\Lambda(x)$ will have high transferability, i.e., perform well on both machine text detection and membership inference. In the remainder of this paper, we analyze a collection of state-of-the-art methods using this framework and measure their degree of transferability in the context of this result.

\subsection{Classifying Methods as Approximate Likelihood Ratio Tests}
Since the true population distribution of human-written text $P_{\mathcal{Q}}$ is inaccessible, methods in both tasks have employed various strategies to approximate this distribution. These strategies can be broadly divided into two approaches:

\paragraph{Approximation via External Reference.} The first approach approximates the true distribution by leveraging an external distribution $P_{\mathcal{M}_\text{ref}}$ as a surrogate for the true population distribution $P_{\mathcal{Q}}$,
\[
P_\mathcal{Q}(x) \approx P_{\mathcal{M}_\text{ref}}(x).
\]
The surrogate distribution $P_{\mathcal{M}_\text{ref}}$ can be approximated via another language model (\textit{Reference} \citep{carlini-etal-2021} from MIAs), a byte-level frequency distribution induced by Huffman encoding (\textit{Zlib} \citep{carlini-etal-2021} from MIAs), a token-level frequency distribution from an external corpus (\textit{DC-PDD} \citep{dc_pdd} from MIAs), or cross-model entropy (\textit{Binoculars} \citep{binoculars} from machine text detection).

\paragraph{Approximation via Text Sampling.}
The second approach leverages text sampling to approximate the true distribution $P_{\mathcal{Q}}$. The likelihood under the true distribution is approximated by the expected likelihood of multiple perturbations $\tilde{x}$ of the target text $x$,
\[
P_\mathcal{Q}(x)  \approx \mathbb{E}_{\tilde{x} \sim \phi(\cdot|x)}[P_\mathcal{M}(\tilde{x}) ].
\]
where $\phi(\cdot|x)$ is a perturbation function that samples variations of the target text $x$. 
This strategy is employed by the \textit{Neighborhood attack} \citep{mattern-etal-2023-membership} from MIAs, \textit{DetectGPT} \citep{pmlr-v202-mitchell23a}, and \textit{Fast-DetectGPT} \citep{bao2023fast} from machine text detection.

Here, the expected likelihood of multiple perturbations can be seen as an approximate marginalization over semantically equivalent paraphrases. When the average delta of likelihood between these paraphrases is larger in $P_\mathcal{M}$ than it is in $P_\mathcal{Q}$, as assumed in \textit{DetectGPT}, applying this marginalization to a particular datapoint would result in a better approximation of the true distribution $Q$.

\paragraph{Implications for Transferability.}
Our theoretical proof shows that the optimal test statistic is identical for membership inference and machine-generated text detection, despite their different objectives. 
Moreover, many methods from both tasks can be interpreted as approximations of the same underlying likelihood ratio, differing in how to approximate the population distribution $P_{\mathcal Q}$. This fundamental connection naturally leads to methods proposed for the two tasks that share similar functional forms. For example, the \textit{Neighborhood attack} for membership inference
\citep{mattern-etal-2023-membership} can be viewed as a finite-sample approximation of \textit{DetectGPT} \citep{pmlr-v202-mitchell23a}, which was introduced for machine-generated text detection. 

Importantly, our theory does not rely on any specific method pair. Instead, it provides a general explanation for why such similarities arise and why methods can transfer across tasks.

In Table \ref{tab:metric_equations}, we provide a breakdown of the methods we test in our work and, where applicable, include a reformulation to fit the categorization we propose. We discuss in more detail in Appendix \ref{app:details_of_methods} the steps we took to get to this general formulation for each example in the author's own notation.

\subsection{Discussion}

While our general formulation covers many methods from both tasks, there are methods that fall outside of our core framing---most notably, the methods that are not ratios but single quantities (e.g., \textit{DC-PDD}, \textit{Min-K\%}). For these methods, we assess transferability empirically and defer a more fully unified theoretical investigation to future work.

In addition, while most methods exhibit a large degree of transferability, there are some methods whose transferability is less strong. Of particular interest is the $\textit{Zlib}$ method which, despite being an approximate likelihood ratio, performs relatively poorly on machine text detection. We discuss some hypotheses for why this is the case in more detail in \S\ref{analyses}.

\section{Quantifying Transferability between MIAs and Machine Text Detection}
\subsection{Experimental Setup}
\label{quantify_transferability}

\paragraph{Membership Inference.}
We evaluate \textbf{MIAs} on the MIMIR dataset \citep{duan2024membership}, which is a large-scale MIA benchmark consisting of \textbf{5 domains}\footnote{To compare the two tasks under a common condition, we focus on textual domains as in machine text detection. See Appendix \ref{full_table_mia_detection} for full results on MIAs, including technical domains: GitHub and DM Mathematics.} included in the Pile \citep{gao2020pile800gbdatasetdiverse}: Wikipedia (knowledge), Pile CC (general web), PubMed Central and ArXiv (academic), HackerNews (dialogue). Members and non-members are sampled from the training and test sets of the Pile, respectively and 13-gram filtering is used to ensure no leakage. We target the \textsc{Pythia} suite: \textbf{5 models} of \textsc{Pythia} \citep{pmlr-v202-biderman23a} with 160M, 1.4B, 2.8B, 6.7B, and 12B parameters.

\paragraph{Machine Text Detection.}
To evaluate the performance of the methods on \textbf{machine text detection}, we use the RAID dataset \citep{dugan-etal-2024-raid}, which is a large-scale detection benchmark consisting of generated text and human-written text in \textbf{8 domains}: Wikipedia and News (knowledge), Abstracts (academic), Recipes (instructions), Reddit (dialogue), Poetry (creative), Books (narrative), Reviews (opinions), and \textbf{5 models}: GPT-2-XL \citep{radford2019language}, MPT-30B-Chat \citep{MosaicML2023Introducing}, LLaMA-2-70B-Chat \citep{touvron2023llama2openfoundation} as open-source models and ChatGPT \citep{chatgpt} and GPT-4 \citep{openai2024gpt4technicalreport} as closed-source models.

\begin{figure}[t]
 \begin{center}
  \centering\includegraphics[width=0.92\columnwidth]{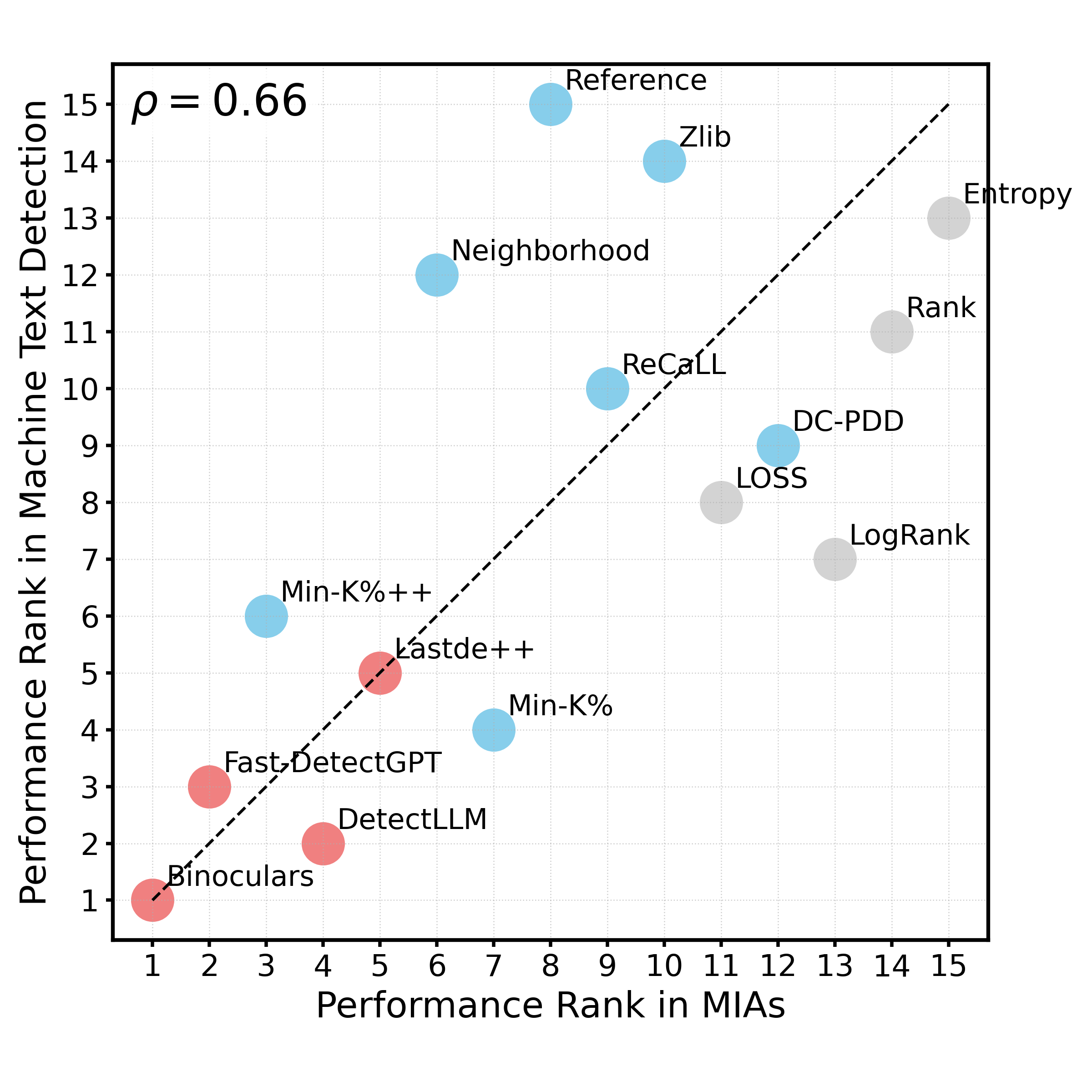}
  \caption{Relationship between method rankings across MIAs and machine text detection. Blue and red plots show \textcolor{skyblue}{MIA methods} and \textcolor{lightcoral}{machine text detectors}. Gray plots indicate \textcolor{gray}{general baselines}. The dashed line denotes equal ranks.}
  \label{rank_correlation_map}
 \end{center}
\end{figure}

\begin{figure*}[t]
 \begin{center}
 \small
  \centering\includegraphics[width=0.98\textwidth]{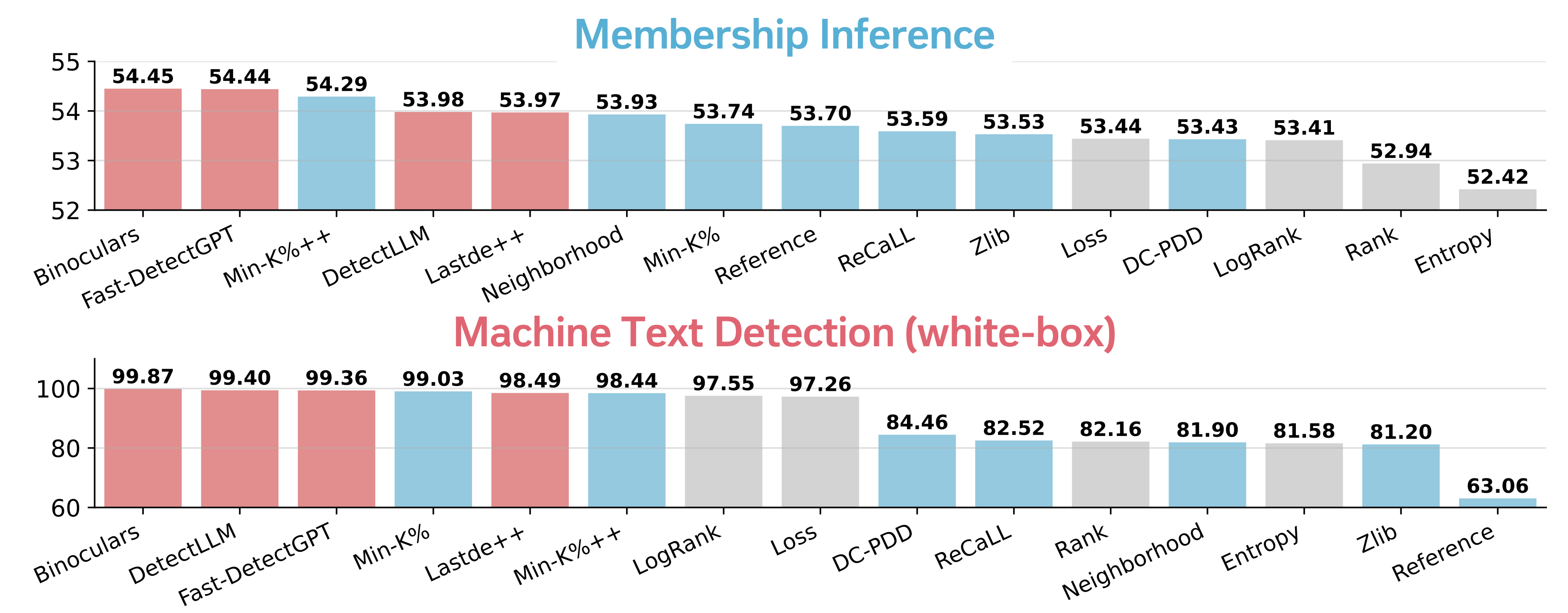}
  \caption{AUROC scores for \textcolor{skyblue}{membership inference attacks} (blue bars) and \textcolor{lightcoral}{generated text detectors} (red bars) across both tasks. \textbf{Top}: Membership inference results on the MIMIR benchmark with 13-gram de-duplication filtering averaged over \textit{five domains} and \textit{five models}. \textbf{Bottom}: Generated text detection results in a white-box setting from the RAID benchmark, averaged over \textit{eight domains} and \textit{three models}. Both MIA methods and machine text detectors have comparable performance in the cross-task setting. Notably, \textit{Binoculars} achieves the best average performance in both tasks. Full results are provided in Appendix \ref{full_table_mia_detection}.}
  \label{detection_cross_performance}
 \end{center}
\end{figure*}


\paragraph{Evaluation Measures.}
To assess the empirical transferability, we compute the performance ranking of all methods on both tasks using the \textbf{AUROC} score and report \textbf{Spearman's rank correlation} coefficient to judge how closely the two rankings match each other\footnote{Our proof simply states that the optimal statistic is identical for both tasks—not that the absolute performance is correlated. Our theoretical contribution is meant to provide intuition as to why the relative rankings of methods are preserved, but it is important to stress that one result does not directly imply the other.}. A high rank correlation implies that the state-of-the-art metrics for one task will perform at or near state-of-the-art on the other. Here, we use AUROC for both tasks, as it reflects a method's overall performance and provides a more reliable measure of transferability than metrics based on a single threshold.

\paragraph{Methods Tested.}
For membership inference, we consider \textbf{7 methods}: \textit{Reference} \citep{carlini-etal-2021}, \textit{Zlib} \citep{carlini-etal-2021}, \textit{Neighborhood attack} \citep{mattern-etal-2023-membership}, \textit{Min-K\% Prob} \citep{min_k}, \textit{Min-K\%++} \citep{min_k_plus_plus}, \textit{ReCaLL} \citep{xie-etal-2024-recall}, and \textit{DC-PDD} \citep{dc_pdd}. For machine text detection, we consider \textbf{5 methods}: \textit{DetectGPT} \citep{pmlr-v202-mitchell23a}, \textit{Fast-DetectGPT} \citep{bao2023fast}, \textit{Binoculars} \citep{binoculars}, \textit{DetectLLM} \citep{su-etal-2023-detectllm}, and \textit{Lastde++} \citep{lastde}. We also compute the \textit{Loss}, \textit{Rank}, \textit{LogRank}, and \textit{Entropy} as general baseline methods. An overview of the methods tested can be found in Table \ref{tab:metric_equations}, more details on each method can be found in Appendix \ref{app:details_of_methods}, and the configurations can be found in Appendix \ref{implementation_details}.

\paragraph{Detection Scenarios.}
For our main rank correlation result, we target a \textbf{white-box setting} for both MIAs and machine text detection, where token probability distributions of target models are accessible.
This allows us to compare the two tasks under a common condition and is consistent with our theoretical formulation, which relies on signals derived from target model probabilities.

To further examine transferability in real-world scenarios, we additionally investigate the \textbf{black-box setting} for machine text detection, targeting closed-source models such as ChatGPT and GPT-4.\footnote{Since the training data of such closed-source models is not accessible, true ground truth for MIAs is inherently not feasible in this setting. Therefore, MIAs are evaluated only in a white-box setting.} 
In this case, where the target model's logits are not available, we employ surrogate models (\textsc{Pythia}-160M \citep{pmlr-v202-biderman23a} and Llama-3-3.2B \citep{grattafiori2024llama3herdmodels}) and report the average detection performance across the surrogates.

\subsection{Main Results}
\label{main_results}
\paragraph{Substantial Rank Correlation Between MIAs and Machine Text Detection.}
\label{results}

Figure \ref{rank_correlation_map} illustrates the relationship between the rankings of all methods when evaluated on MIAs and on machine text detection.
The rankings are based on average performance on MIAs (across five domains and five target models) and machine text detection (across eight domains and three generators). 
We compute the rank correlation over all 15 methods and obtain a statistically significant Spearman's correlation of $\rho = 0.66$ ($p<0.01$).
This result indicates that many methods originally proposed for MIAs also perform well in machine text detection, and vice versa. 
Notably, when focusing on stronger methods (top-10 on MIAs), we observe an even stronger correlation of $\rho = 0.78$ ($p<0.01$). These findings align with our theoretical framework, suggesting that methods achieving strong performance on one task, by better approximating the shared optimal metric, tend to exhibit higher transferability and perform well on the other task.

\paragraph{Superior Performance from the Other Task.}
Figure \ref{detection_cross_performance} shows the AUROC performance of all methods from both MIAs and machine text detection evaluated on MIAs, and their performance evaluated on machine text detection. 
The results on MIAs are averaged across five domains and five target models, while those on machine text detection are averaged across eight domains and three generators. 
Remarkably, we observe that \textit{Binoculars}, originally proposed as a machine text detector, achieves the best average performance in both MIAs and machine text detection. 
This result suggests that current evaluations in MIAs may be biased by overlooking stronger methods from machine text detection, potentially leading to conclusions that miss valuable insights. 
Conversely, in machine text detection, methods from MIAs already demonstrate competitive or even superior performance. These findings call for greater cross-task awareness and development.

\begin{figure*}
 \begin{center}
 \small
 \centering
 \includegraphics[width=0.98\textwidth]{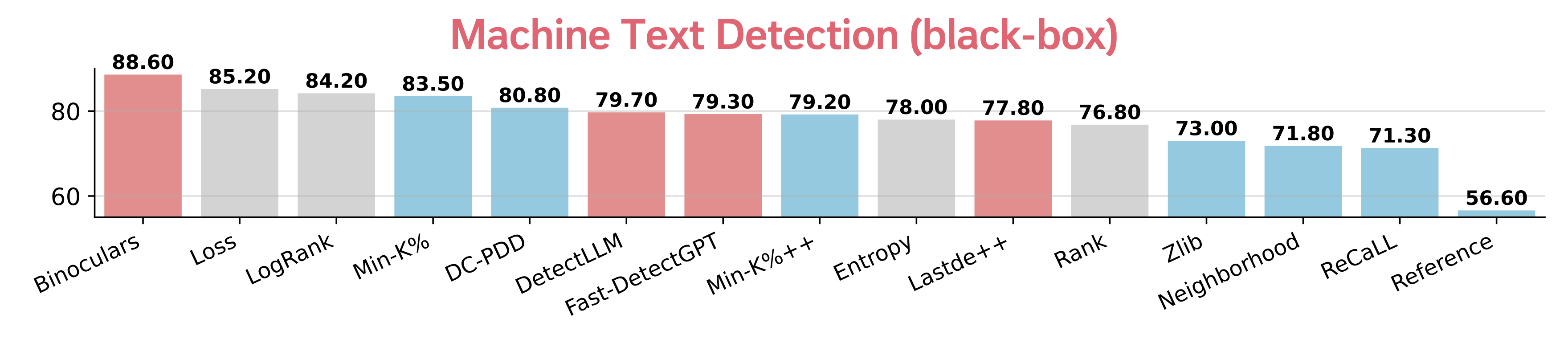}
  \caption{AUROC scores for \textcolor{skyblue}{membership inference attacks} (blue bars) and \textcolor{lightcoral}{generated text detectors} (red bars) on generated text detection in a black-box setting from the RAID benchmark, averaged over \textit{eight domains} and \textit{two closed-source models}. Even in this real-world scenario, MIA methods such as \textit{Min-K\%} and \textit{DC-PDD} achieve performance on par with or better than strong detectors. Full results are provided in Appendix \ref{full_table_mia_detection}.
  }
  \label{black_box_detection}
 \end{center}
\end{figure*}

\subsection{Analysis}
\label{analyses}
\paragraph{Transferability in Real-World Scenarios.}
We further assess transferability in real-world scenarios by evaluating how well MIA methods perform in a black-box setting of machine text detection on texts generated by ChatGPT and GPT-4. Figure \ref{black_box_detection} shows the average AUROC performance of all methods from MIAs and machine text detection across eight domains and the two generators. While \textit{Binoculars} still outperforms other methods by a large margin, MIA methods such as \textit{Min-K\%} and \textit{DC-PDD} achieve performance on par with or better than strong detectors.
These results provide promising evidence of the transferability of MIAs to machine text detection in real-world scenarios.

\paragraph{Similar Prediction Score Distributions across Methods.}
To further illustrate the transferability between MIAs and machine text detection, we compare the prediction score distributions of an MIA method and a machine text detector when applied to the same task. Figure \ref{cross_task_distribution_plot} presents the distributions of \textit{Min-K\%++} and \textit{Binoculars}, which demonstrate strong cross-task performance, on MIAs and machine text detection. The domain in both tasks is Wikipedia, and the target model or generator is \textsc{Pythia}-12B. 
To quantify the similarity in distributional shape, we compute the Jensen–Shannon distance between the prediction score distributions produced by \textit{Min-K\%++} and \textit{Binoculars} within each task: 0.14 for MIAs and 0.11 for machine text detection.
These small distances indicate that \textit{Min-K\%++} and \textit{Binoculars} produce closely aligned prediction score distributions within both tasks, providing further evidence of their transferability.

\paragraph{\textit{Zlib} as an Outlier Illustrates a Task Difference: Different Prior Distributions.}
We take \textit{Zlib} as an example of limited transferability between MIAs and machine text detection. \textit{Zlib} ranks 10th out of 15 methods in MIAs but drops to 14th in detection, where it calibrates the loss by dividing it by the zlib compression entropy.

\begin{figure*}
 \begin{center}
 \small
 \centering
 \includegraphics[width=0.98\textwidth]{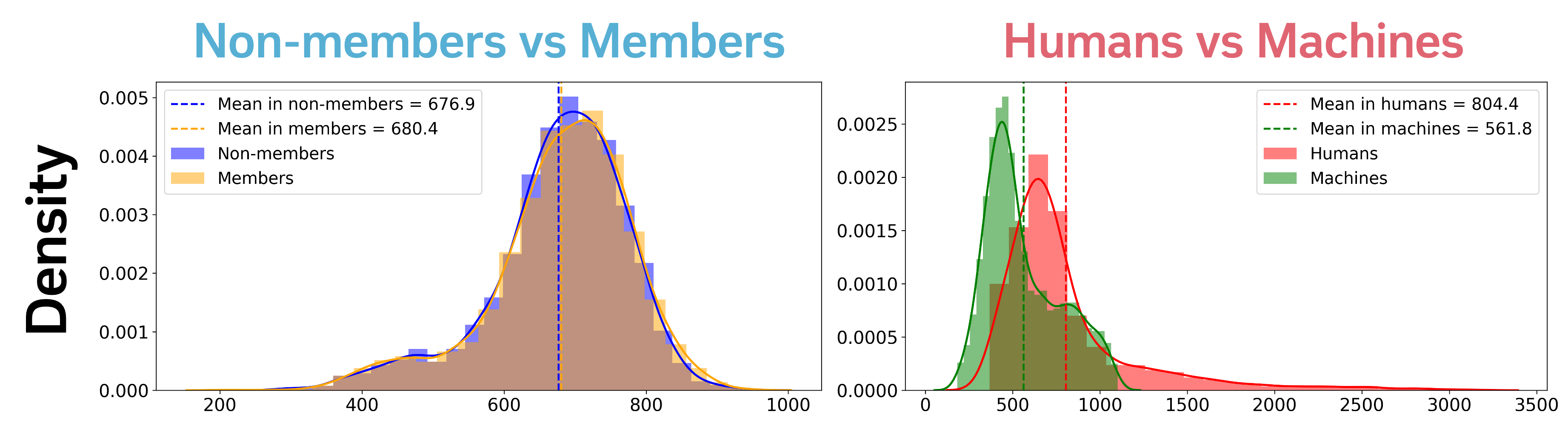}
  \caption{\textbf{Zlib compression entropy distribution} in \textcolor{skyblue}{membership inference} (non-members vs. members) and \textcolor{lightcoral}{machine text detection} (humans vs. machines), averaged over 3,000 randomly sampled texts for each dataset. The entropy converges between classes in MIAs but diverges in detection with a long tail, due to different prior distributions.}
  \label{zlib_distribution_plot}
 \end{center}
\end{figure*}

In MIAs, both classes are drawn from the same human-written text distribution, whereas this is not the case for machine text detection.
Machine-generated texts are known to be more compressible than human-written ones \citep{tulchinskii2023intrinsic,mao2025losslesscompressionlargelanguage}, and we also find the trend in our setting. 
As shown in Figure \ref{zlib_distribution_plot}, the zlib entropy converges between classes in MIAs but diverges in machine text detection, based on 3,000 randomly sampled texts from all domains and models in each dataset. 
In the detection, both the loss and the zlib entropy shift in the same direction between human-written and machine-generated texts, yielding similar \textit{Zlib} scores across classes and little discriminative signal.
Consequently, while moderately effective for MIAs, \textit{Zlib} transfers poorly to machine text detection, highlighting a key task difference: \textbf{different prior distributions}.
We leave a comprehensive analysis of other methods with limited transferability for future work.

\section{Related Work}
\paragraph{Proximity between MIAs and Machine Text Detection.}
Previous work has only briefly noted similarities between a specific method pair in MIAs and machine text detection. Specifically, \citet{min_k} mentions that the \textit{Neighborhood attack} \citep{mattern-etal-2023-membership} from MIAs and \textit{DetectGPT} \citep{pmlr-v202-mitchell23a} from machine text detection share a similar formulation, as both estimate the local curvature of model likelihoods via text perturbations. More recently, \citet{naseh2025syntheticdatamisleadevaluations} reported that current MIAs often misclassify machine-generated non-members as members. 

In contrast to prior studies that are limited to a single method pair or task-specific observations, we frame transferability as a general research question. We conduct the first comprehensive study of transferability, introducing a unified theoretical formulation and validating it empirically through large-scale experiments.

\paragraph{Optimality of Likelihood Ratio Tests for MIAs.}
Foundational work by \cite{carlini-etal-2022} was the first to utilize the optimality of the likelihood ratio test as a starting point from which to derive membership inference attacks. They propose an online likelihood ratio attack (LiRA) where they train sets of ``shadow'' models on random samples of the data distribution with and without the target point $x$. They then fit two Gaussians to the confidences of the ``in'' and ``out'' models to approximate a parametric likelihood ratio test. They show that these tests achieve strong performance at low false positive rates. While their specific technique is not as practical in the context of multi-billion parameter foundation models with trillion-token pretraining datasets, their work nonetheless provides solid empirical justification for the usage of likelihood ratio tests as a framework from which to reason about strong MIA performance.

\paragraph{Membership Inference.}
Membership inference is a task to determine whether a given sample was part of a given model's training data \citep{shokri}. In the context of language models, recent studies have applied MIAs to pre-training data detection. Since language models tend to show higher likelihood on members compared to non-members, such studies leverage statistical features of a target model, such as likelihood \citep{carlini-etal-2021}, likelihood calibrated by another language model \citep{carlini-etal-2021}, negative log probability curvature \citep{mattern-etal-2023-membership}, and average log-likelihood of the \textit{k}\% tokens with lowest probabilities \citep{min_k}. Other MIA methods are detailed in Appendix \ref{overview_mia}. 

\paragraph{Machine-generated Text Detection.}
Many studies on machine text detection have investigated supervised methods, including training a classifier with human-written and machine-generated texts with labels. Supervised classifiers aim to capture stylistic or semantic differences between human-written and machine-generated texts, with simple n-gram features \citep{ippolito-etal-2020-automatic,crothers2023machinegeneratedtextcomprehensive,verma-etal-2024-ghostbuster} or neural representations \citep{li-etal-2024-mage,wang-etal-2024-m4gt}. 
Since MIAs fundamentally examine how much a target model memorizes a text by exploiting signals from the target model such as likelihood, our investigation of the transferability focuses on zero-shot detectors that similarly rely on statistical features from the model, rather than supervised classifiers that do not.
Zero-shot detectors leverage statistical features, including entropy \citep{ent08}, likelihood \citep{solaiman2019releasestrategiessocialimpacts}, negative curvature of log probabilities \citep{pmlr-v202-mitchell23a}, and cross-model entropy \citep{binoculars}. Other zero-shot detectors are detailed in Appendix \ref{overview_detection}.

\section{Conclusion}
We comprehensively study the transferability between membership inference attacks (MIAs) and machine-generated text detection. Our theoretical and empirical investigations reveal that 1) Many methods from both tasks that exhibit high transferability can be reduced to a general formulation that measures the discrepancy between the likelihood of a text under a target model and under the true distribution, reflecting their shared objective to approximate the true distribution and contributing to the transferability and 2) Many methods originally proposed for MIAs perform well in machine text detection, and vice versa, as evidenced by substantial rank correlation across the tasks and 3) Notably, a method originally designed for machine text detection surpasses state-of-the-art MIA methods on MIAs, demonstrating the practical impact of the transferability.
These findings call for greater cross-task awareness, closer collaboration, and fair evaluation across the two research communities.

\section*{Author Contributions}
Ryuto Koike and Liam Dugan co-designed the project and jointly developed its conceptual framing. Ryuto Koike wrote the code, ran the experiments, and drafted the initial version of the paper. Liam Dugan proved the theoretical formulation and contributed to writing and revising the paper. Masahiro Kaneko, Chris Callison-Burch, and Naoaki Okazaki supervised the project and provided valuable feedback and helpful edits to the draft.

\section*{Acknowledgements}
We thank Shannon Sequiera for helping to check the mathematical derivations and helping improve the clarity of the notation in the general formulation. We are also grateful to Andrew Zhu, Alyssa Hwang, and Youmi Ma for their feedback on early versions of the manuscript.

This research was developed with funding from the Defense Advanced Research Projects Agency’s (DARPA) SciFy program (Agreement No. HR00112520300) as well as the Office of the Director of National Intelligence (ODNI), Intelligence Advanced Research Projects Activity (IARPA), via the HIATUS Program contract \#2022-22072200005. The views expressed are those of the author and do not reflect the official policies, either expressed or implied, of ODNI, IARPA, DARPA, or the U.S. Government. The U.S. Government is authorized to reproduce and distribute reprints for governmental purposes notwithstanding any copyright annotation therein.
These research results were also obtained from the commissioned research (No.22501) by National Institute of Information and Communications Technology (NICT), Japan. In addition, this work was supported by JST SPRING, Japan Grant Number JPMJSP2106.

\nocite{langley00}

\bibliography{example_paper}
\bibliographystyle{icml2026}

\newpage
\appendix
\onecolumn
\section{Proof that $s = P_{\mathcal{M}}(x)$ is sufficient}
\label{app:detailed_proof}
In our proof of Theorem \ref{unified_framework} we transform the likelihood ratio test for the membership inference task as follows:
\[H_0: \mathcal{M} \leftarrow \mathcal{T}(\mathcal{D}'),\; H_1: \mathcal{M} \leftarrow \mathcal{T}(\mathcal{D}'\cup\{x\}).\]\[\Lambda_{\text{MI}}(x) = \frac{P(M=\mathcal{M} | H_1)}{P(M=\mathcal{M} | H_0)} = \frac{P(s = P_\mathcal{M}(x) | H_1)}{P(s = P_\mathcal{M}(x)| H_0)}\]
where $s = P_M(x)$ is a random variable representing the distribution of the likelihood of $x$ under $M$. For this to be valid, we must show that $s$ is sufficient for $M$. In other words, we must show that
\[P(M \mid s, H_0) = P(M \mid s, H_1)\]
Let $\mathcal{M}$ be a model following our asymptotic assumptions laid out in Theorem \ref{unified_framework} and let $N = |\mathcal{D}_\text{train}|$ be the size of its training set. The likelihood of each text $y \in \mathcal{D}_\text{train}$ is defined as \[P_\mathcal{M}(y) = \dfrac{1}{N}\text{count}(y,\mathcal{D}_\text{train})\]
where $\text{count}(y,\mathcal{D}_\text{train})$ is the number of times a particular text $y$ occurs in $\mathcal{D}_\text{train}$. Let $\hat{\textit{\textbf{p}}}$ be the vector of likelihoods $P_{M}(y)$ for all $y \in \mathcal{D}_\text{train}$. This vector contains all information needed to reproduce the full output distribution of $\mathcal{M}$.\footnote{Note here that we assume models $M_1$ and $M_2$ are equivalent if $P_{M_0}(x) = P_{M_1}(x) \;\; \forall \;\; x \in \mathcal{X}$.} Thus, $\hat{\textit{\textbf{p}}}$ is sufficient for $M$ and
$P(M \mid \hat{\textit{\textbf{p}}}, H_0) = P(M \mid \hat{\textit{\textbf{p}}}, H_1)$.

Now, let $\hat{\textit{\textbf{p}}}_{\neg x}$ be the vector of likelihoods for all elements not equal to $x$. In order to show that $s$ is sufficient, we must show that $\hat{\textit{\textbf{p}}}_{\neg x}$ is distributed identically regardless of the hypothesis. In other words we want to show that:
\[P_{M}(y \mid s, H_0) = P_{M}(y \mid s, H_1) \;\; \forall \;\; y \neq x, \; y \in \mathcal{D}_\text{train}\]
To do this, we first note the difference between our two hypotheses: For $H_0$, $\mathcal{D}_\text{train}$ was constructed by taking $N$ uniform draws from $P_{Q}$ and, for $H_1$, $\mathcal{D}_\text{train}$ was constructed by taking $N - 1$ uniform draws from $P_{Q}$ and one fixed draw of $x$. Let $N - \text{count}(x, \mathcal{D}_\text{train})$ be the total number of draws in $\mathcal{D}_\text{train}$ allocated to texts that are not $x$. We note that, regardless of whether we are in $H_0$ or $H_1$, these draws are distributed identically. Thus, for any $y \neq x$ we can model the frequency that $y$ appears in $\mathcal{D}_\text{train}$ via the binomial
\[\text{count}(y, \mathcal{D}_\text{train}) \sim \text{Binomial}(N - \text{count}(x, \mathcal{D}_\text{train}), P_{Q})\]
Substituting in likelihoods we get that:
\[P_{\mathcal{M}}(y) \sim \frac{1}{N} \text{Binomial}(N - N\cdot P_{\mathcal{M}}(x), P_{Q})\]

This means that, conditioned on fixed knowledge of $s = P_\mathcal{M}(x)$, we get that:
\[P_{\mathcal{M}}(y \mid s, H_0) = P_{\mathcal{M}}(y \mid s, H_1) \;\; \forall \;\; y \neq x, \; y \in \mathcal{D}_\text{train}\]
which implies that $P(M \mid s, H_0) = P(M \mid s, H_1)$ and thus $s$ is a sufficient statistic for our hypothesis test.

\section{Details of Methods}
\label{app:details_of_methods}
\subsection{Baselines}
(1) \textbf{\textit{Loss}} simply uses the target sample $x$'s loss against the model $\mathcal{M}$: $f(\mathbf{x};\mathcal{M}) = \mathcal{L}(\mathbf{x};\mathcal{M})$. The hypothesis is that members and machine-generated texts will have a higher likelihood on average than non-members and human-written texts.

(2) \textbf{\textit{Entropy}} measures the expected likelihood of the next token given the preceding tokens at each time step under the model's distribution. $f(\mathbf{x};\mathcal{M}) = \mathbb{E}_{\tilde{x}\sim\mathcal{M}}[\mathcal{L}(\tilde{x};\mathcal{M})]$. The hypothesis is that machine-generated texts and member texts will have low entropy as the model ``understands'' the context more and can better model the next token distribution.

(3) \textbf{\textit{Rank}} measures the average rank of the next token in the model's probability distribution at each time step. $f(x;\mathcal{M}) = \frac{1}{n}\sum_{i=1}^{n}\mathrm{Rank}(x_i;\mathcal{M})$. The hypothesis is that generated text and member text will have a higher average rank than human-written text.

(4) \textbf{\textit{LogRank}} measures the average log rank of the next token: $f(x;\mathcal{M}) = \frac{1}{n}\sum_{i=1}^{n}\log(\mathrm{Rank}(x_i;\mathcal{M}))$. This metric smooths out contributions from very low rank tokens to reflect the probabilistic nature of rank information. Similarly to the previous metric, the hypothesis is that generated text and member text have a high average log rank.

\subsection{Membership Inference Attacks}
\label{overview_mia}

(1) \textbf{\textit{Reference}} \citep{carlini-etal-2021} uses the difference in the target sample $x$'s loss between the model $\mathcal{M}$ and another reference model $\mathcal{M}_\text{ref}$. We follow the original author's implementation by taking a smaller-size reference model for each of the models we tested: $f(\mathbf{x};\mathcal{M}) = \mathcal{L}(\mathbf{x};\mathcal{M}) - \mathcal{L}(\mathbf{x};\mathcal{M}_\text{ref})$. In our notation, this can be represented as $-\log (P_{\mathcal{M}}(x) / P_{\mathcal{M}_\text{ref}}(x))$. This method falls into the external reference category of likelihood ratio approximation.

(2) \textbf{\textit{Zlib}} \citep{carlini-etal-2021} employs the ratio of $\mathcal{L}(\mathbf{x};\mathcal{M})$ and the zlib compression score of the target sample $x$: $f(\mathbf{x};\mathcal{M}) = \mathcal{L}(\mathbf{x};\mathcal{M}) \  / \ {\rm zlib}(\mathbf{x})$. The zlib compression score is computed by constructing a dictionary of repeated substrings from the text and encoding each dictionary entry into a string of bits using Huffman Coding \citep{huffman1952method}. The compression rate is thus a representation of the entropy of the empirical substring distribution of the text. Thus the zlib method can be equivalently rewritten as $f(\mathbf{x};\mathcal{M}) = - \log P_\mathcal{M}(\mathbf{x}) \  / \ \mathbb{E}_{x\sim \mathcal{M}_\text{ref}}[- \log P_\mathcal{M_\text{ref}}(\mathbf{x})]$ where $\mathcal{M}_\text{ref}$ represents that empirical substring distribution. We see here that this method falls into the external reference category of likelihood ratio approximation.

(3) \textbf{\textit{Neighborhood attack}} \citep{mattern-etal-2023-membership} compares $\mathcal{L}(\mathbf{x};\mathcal{M})$ to the average loss of \textit{neighborhood samples} $\tilde{\mathbf{x}}$, which are samples crafted by perturbing the target sample $x$: $f(\mathbf{x};\mathcal{M}) = \mathcal{L}(\mathbf{x}; \mathcal{M}) - \frac{1}{n} \sum_{i=1}^n \mathcal{L}(\tilde{\mathbf{x}}^{(i)}; \mathcal{M})$. It hypothesizes that members will show a lower loss compared to their neighborhood samples. Since the quantity $\mathbb{E}_{\tilde{x}\sim\phi(x)}[\log (P_{\mathcal{M}}(x) / P_{\mathcal{M}}(\tilde{x}))]$, we consider this method to be in the text sampling category of likelihood ratio approximation.

(4) \textbf{\textit{Min-K\%}} \citep{min_k} calculates the average of log-likelihood of the $k$\% tokens with lowest probabilities: $f(\mathbf{x};\mathcal{M}) = \frac{1}{k} \sum_{x_i \in \text{min-}k(\mathbf{x})} {\rm log}\ p(x_i \ | \ x_{<i}; \mathcal{M})$. The intuition is that members will include fewer outlier tokens with low probability compared to non-members.

(5) \textbf{\textit{Min-K\%++}} \citep{min_k_plus_plus} computes the average of the log-likelihood of the $k$\% tokens with lowest probabilities, where each value is standardized over the model's vocabulary:
$f(\mathbf{x};\mathcal{M}) = \frac{1}{k} \sum_{x_i \in \text{min-k\%}} ({\rm log}\ p(x_i \mid x_{<i}; \mathcal{M}) - \mu_{x_{<i}}) / \sigma_{x_{<i}}$. In our notation, this can be written as, $\frac{1}{k}\sum_{i\in \text{min-}k\%}
 \Phi(-\log P_{\mathcal{M}}(x_i) + \mathbb{E}_{\tilde{x}_i\sim\mathcal{M}}[\log P_{\mathcal{M}}(\tilde{x}_i)])$, for brevity in our Table \ref{tab:metric_equations}, we use $\Phi(x) = x / \sigma_x$ as shorthand for the normalization by the standard deviation. Intuitively, this metric measures how unexpectedly surprising a particular token is over the vocabulary distribution.

(6) \textbf{\textit{ReCaLL}} \citep{xie-etal-2024-recall} computes the relative conditional log-likelihood between $x$ and $P \oplus x$ where $P$ is a set of non-member examples $P = p_1 \oplus \cdots \oplus p_n$. It hypothesizes that non-members will have lower log-likelihoods than members, given a non-member context. We represent this in our table as $\mathbb{E}_{\tilde{x}\sim \phi(x)}[\log P_{\mathcal{M}}(\tilde{x})] / (\log P_{\mathcal{M}}(x))$ where $\phi(x) = P \oplus x$. We consider this method to be in the likelihood ratio by text sampling category.

(7) \textbf{\textit{DC-PDD}} \citep{dc_pdd} computes the cross-entropy between the token likelihoods under the model $\mathcal{M}$ and the empirical laplace-smoothed unigram token frequency distribution under some reference corpus $\mathcal{D}'$. In the authors' notation this is $f(\mathbf{x};\mathcal{M}) = -\frac{1}{n} \sum_{i=1}^n p(x_i; \mathcal{M}) \cdot {\rm log}\ p(x_i; \mathcal{D}')$ where $p(x_i;\mathcal{D}') = \frac{\text{count}(x_i) + 1}{N' + |V|}$. In our table, we represent this metric equivalently as $\mathbb{E}_{\tilde{x}\sim\mathcal{M}}[-\log P_{\mathcal{M}_\text{ref}}(\tilde{x})]$ where $\mathcal{M}_\text{ref}$ represents the unigram token frequency distribution under $\mathcal{D}'$.

\subsection{Machine-generated Text Detectors}
\label{overview_detection}

(1) \textbf{\textit{DetectGPT}} \citep{pmlr-v202-mitchell23a} computes the degree to which the log-likelihood function under the suspected model has negative curvature for the given target text. They do this by perturbing the sequence using a T5 model \cite{raffel2023exploringlimitstransferlearning} and evaluating the change in probability. The functional form looks like $f(x;\mathcal{M}) = \mathbb{E}_{\tilde{x}\sim\phi(x)}[\log (P_{\mathcal{M}}(x) / P_{\mathcal{M}}(\tilde{x}))]$. We directly lift this notation for use in our Table \ref{tab:metric_equations}. This metric is an example of a text sampling based likelihood ratio approximation.

(2) \textbf{\textit{Fast-DetectGPT}} \citep{bao2023fast} forgoes the expensive perturbation approach used by DetectGPT in favor of using the token likelihoods of the perturbation model directly to compute the expectation. They also divide by the mean of sample variances to further smooth out the metric. The full formulation is
\[
f(x;\mathcal{M}) = \frac{\log p(x;\mathcal{M}) - \mathbb{E}_{\tilde{x}\sim q(x)}[\log p(\tilde{x};\mathcal{M})]}{\sqrt{\mathbb{E}_{\tilde{x}\sim{q(x)}}[(\log p(\tilde{x};\mathcal{M}) - \mathbb{E}_{\tilde{x}\sim q(x)}[\log p(\tilde{x};\mathcal{M})])^2]}}
\]
We report this using the shorthand $\Phi(x) = x / \sigma_x$ in our Table \ref{tab:metric_equations} to represent this quantity. We consider this metric another example of the text sampling based likelihood ratio approach. 

(3) \textbf{\textit{Binoculars}} \citep{binoculars} computes the ratio of the perplexity to the cross entropy of the text under some reference model $\mathcal{M}_\text{ref}$. The full formulation using the authors' notation is as follows:
\[B_{\mathcal{M}_1, \mathcal{M}_2}(s) = 
\frac{\sum_{i=1}^{L}\log \left( \mathcal{M}_1(s)_i \right)}{\sum_{i=1}^{L}\mathcal{M}_1(s)_i \cdot \log \left( \mathcal{M}_2(s)_i \right)}\]
We re-formulate this objective equivalently in our own notation as:
\[
(\log P_{\mathcal{M}}(x)) / \mathbb{E}_{\tilde{x}\sim\mathcal{M}}[\log P_{\mathcal{M}_\text{ref}}(\tilde{x})]
\]
The intuition behind this metric is that it measures how much more likely a given text is than what we would expect according to some reference model. We consider this metric a case of approximation via reference model.

(4) \textbf{\textit{DetectLLM}} \citep{su-etal-2023-detectllm} is a variant of DetectGPT that uses the log rank as the core quantity to test rather than the log likelihood. The authors propose two metrics, Log-Likelihood Log-Rank Ratio (LRR) and Normalized Log-Rank Perturbation (NPR). For the purposes of our work we consider the NPR metric which has superior performance and is characterized by the following formulation:
\[
\text{NPR} = \frac{\frac{1}{n}\sum_{p=1}^n \log r_{\theta}(\tilde{x}_p)}{\log r_{\theta}(x)}
\]
where $r_{\theta}(x_i)$ represents the rank of the token $x$ in the model's output distribution. We reformulate this equivalently as
\[
\frac{\mathbb{E}_{\tilde{x}\sim \phi(x)}[\mathcal{R}(\tilde{x};\mathcal{M})]}{\mathcal{R}(x;\mathcal{M})}
\]
using the notation $\mathcal{R}(x;\mathcal{M})$ to denote the log rank similar to how $\mathcal{L}(x;\mathcal{M})$ denotes the negative log likelihood. While this metric doesn't quite fall cleanly into our likelihood ratio categorization (since it does not compute likelihood) we nonetheless note the strong similarities between this metric and other approximate likelihood ratios. 

(5) \textbf{\textit{Lastde++}} \citep{lastde} utilizes a quantity known as multi-scale diversity entropy (MDE) \citep{multiscale-diversity-entropy} to measure the local fluctuations in likelihood across a particular text sequence. Their metric is:
\[
\mathrm{Lastde}(x;\mathcal{M}) = \frac{\mathcal{L}(x;\mathcal{M})}{\mathrm{StdDev}(\{\mathrm{DE}(s, \varepsilon, 1), ..., \mathrm{DE}(s, \varepsilon, \tau')\})}
\]
\[
\mathrm{DE}(s, \varepsilon, \tau) = -\frac{1}{\ln{\varepsilon}}\sum_{i=1}^{\varepsilon}P_i^{(\tau)}\ln P_i^{(\tau)}
\]
Where $P_i^{(\tau)}$ measures the \textit{diversity} of text, i.e. the extent to which adjacent segments of tokens have similar probability sequences (see \cite{lastde}). Intuitively the Lastde metric can be thought of as comparing likelihood to the expected diversity. Lastde++ takes this quantity and applies a sampling based perturbation and normalization similar to FastDetectGPT \citep{bao2023fast}. 
\[
\mathrm{Lastde}\text{++}(x) = \frac{\mathrm{Lastde}(x;\mathcal{M}) - \mathbb{E}_{\tilde{x}\sim \phi(x)}[\mathrm{Lastde}(\tilde{x},\mathcal{M})]}{\sqrt{\mathbb{E}_{\tilde{x}\sim{\phi(x)}}[(\mathrm{Lastde}(\tilde{x},\mathcal{M}) - \mathbb{E}_{\tilde{x}\sim \phi(x)}[\mathrm{Lastde}(\tilde{x},\mathcal{M})])^2]}}
\]
This can be thought of as measuring whether or not the quantity tracked by $\mathrm{Lastde}(x)$ is at a local maximum for the particular sequence of text.

We consider neither Lastde++ or Lastde to be approximating likelihood ratios as the diversity-entropy ($\mathrm{DE}$) metric seems to be more intuitively thought of as measuring variance rather than likelihood. We leave to future work a more thorough analysis of these variance-based metrics.

\section{Implementation Details}
\label{implementation_details}
\subsection{Membership Inference Attacks}
\paragraph{\textit{Reference}} \citep{carlini-etal-2021} Following the original author's implementation, we take a smaller-size reference model for each target model. In MIAs, we use \textsc{Pythia}-70M for all target models. In white-box machine text detection, we use GPT2-Small as the smaller model for GPT2-XL, MPT-7B-Chat for MPT-30B-Chat, and LLaMA-7B-Chat for LLaMA-70B-Chat.
In black-box machine text detection, we use LLaMA-3-1B as the smaller model for LLaMA-3-3.2B and \textsc{Pythia}-70M for \textsc{Pythia}-160M.

\paragraph{\textit{Neighborhood attack}} \citep{mattern-etal-2023-membership} We use the repository\footnote{\url{https://github.com/mireshghallah/neighborhood-curvature-mia}} default settings, namely T5-Large as mask filling model and 0.3 as the masking rate, across MIAs, white-box machine text detection, and black-box machine text detection. 

\paragraph{\textit{Min-K\%}} \citep{min_k} We use the setting that was found to ensure the favorable performance in the original paper, namely $k=20$ as the percent of tokens with the lowest probabilities. Likewise, the token percentage $k$ for Min-K\%++ \citep{min_k_plus_plus} is also set as $k=20$ for fair comparison.

\paragraph{\textit{ReCaLL}} \citep{xie-etal-2024-recall} 
We set the number of prefixes to $n=10$, which has been shown in the original paper to yield favorable performance. In MIAs, for each domain, we retrieve non-members as prefixes from the Pile dataset, excluding those in the MIMIR benchmark. To adapt ReCaLL to machine text detection, we use human-written texts as prefixes, since they belong to the negative class. For each domain, these human-written prefixes are retrieved from a subset of the RAID benchmark that was not used in our test set.

\paragraph{\textit{DC-PDD}} \citep{dc_pdd} Following the official GitHub repository\footnote{\url{https://github.com/zhang-wei-chao/DC-PDD}}, we take a subset of C4 \citep{raffel2023exploringlimitstransferlearning} and build a token frequency distribution with the tokenizer of each target model in MIAs and white-box machine text detection, and of each surrogate model in black-box detection.

\subsection{Machine Text Detectors}
\paragraph{\textit{DetectGPT}} \citep{pmlr-v202-mitchell23a} We follow the repository\footnote{\url{https://github.com/eric-mitchell/detect-gpt}} default settings, namely T5-Large as mask filling model and 0.3 as the masking rate, across MIAs, white-box machine text detection, and black-box machine text detection. 

\paragraph{\textit{Fast-DetectGPT}} \citep{bao2023fast} Following the original paper, we employ the target model as both the scoring and perturbation model in MIAs and white-box machine text detection. In black-box detection, we instead use surrogate models as those (see \S\ref{quantify_transferability}).

\paragraph{\textit{Binoculars}} \citep{binoculars} is reported to work best when the target and reference models are similar in performance. 
Following the original setting, we use the official code\footnote{\url{https://github.com/ahans30/Binoculars}} to compute perplexity under a reference model. For MIAs, since the \textsc{Pythia} does not provide corresponding chat versions for each model size, we use the \textsc{Pythia}-deduped models as references.
In white-box detection, we use GPT2-XL-Chat \footnote{\url{lgaalves/gpt2-xl_lima}} for GPT2-XL, MPT-30B for MPT-30B-Chat, and LLaMA-2-70B for LLaMA-2-70B-Chat. In black-box detection, we adopt LLaMA-3.2-3B-instruct for LLaMA-3-3.2B and \textsc{Pythia}-160M-deduped for \textsc{Pythia}-160M.

\paragraph{\textit{DetectLLM}} \citep{pmlr-v202-mitchell23a} In line with Fast-DetectGPT, we utilize the target model as both the scoring and perturbation model in MIAs and white-box detection. For black-box detection, we instead use surrogate models (see \S\ref{quantify_transferability}).

\paragraph{\textit{Lastde++}} \citep{lastde} In our implementation, we use the official code and use the repository\footnote{\url{https://github.com/TrustMedia-zju/Lastde_Detector}} default settings of the sliding window size $s=4$, the interval precision $\epsilon=8$, and the number of scales $\tau'=15$.

\section{Full Performance on Membership Inference and Machine Text Detection}
\label{full_table_mia_detection}
Table \ref{mia_detection_performance_result_mimir}, Table \ref{white_box_llm_detection_performance_result}, and Table \ref{black_box_llm_detection_performance_result} report the full results on membership inference, white-box machine text detection, and black-box machine text detection, respectively.

\begin{table*}[h]
\caption{Full comparison of membership inference performances (AUROC) of MIA methods and machine text detectors on the MIMIR benchmark with 13-gram deduplication. Target models are \textsc{Pythia} with 160M, 1.4B, 2.8B, 6.9B, and 12B parameters. Gray, blue, and red areas indicate general baseline methods, MIA methods, and machine text detectors, respectively. Textual domains are the bolded columns (Wikipedia, Pile CC, PubMed, ArXiv, HackerNews).}
\centering
\fontsize{7.5pt}{7.5pt}\selectfont
\setlength{\tabcolsep}{1.2pt} 
\renewcommand{\arraystretch}{1.3} 
\begin{tabular}{lcccccccccccccccccccc}
\toprule
 \multicolumn{1}{c}{\multirow{2.5}{*}{\textbf{Method}}} & \multicolumn{5}{c}{\textbf{Wikipedia}} & \multicolumn{5}{c}{GitHub} & \multicolumn{5}{c}{\textbf{Pile CC}} & \multicolumn{5}{c}{\textbf{PubMed Central}} \\
\cmidrule(lr){2-6} \cmidrule(lr){7-11} \cmidrule(lr){12-16} \cmidrule(lr){17-21} 
 & 160M & 1.4B & 2.8B & 6.9B & 12B & 160M & 1.4B & 2.8B & 6.9B & 12B & 160M & 1.4B & 2.8B & 6.9B & 12B & 160M & 1.4B & 2.8B & 6.9B & 12B \\
\midrule
\rowcolor{gray!30}
Loss & 51.2 & 53.4 & 54.1 & 55.6 & 56.5 & 76.3 & 80.2 & 81.4 & 82.7 & 83.6 & 50.1 & 51.0 & 51.2 & 52.1 & 52.7 & 50.9 & 52.1 & 52.7 & 53.4 & 54.0 \\
\rowcolor{gray!30}
Rank & 49.7 & 52.9 & 53.9 & 56.2 & 57.5 & 70.0 & 74.6 & 75.7 & 77.4 & 77.8 & 50.9 & 51.6 & 51.8 & 52.2 & 52.1 & 51.9 & 52.3 & 52.4 & 53.0 & 53.6 \\
\rowcolor{gray!30}
LogRank & 51.3 & 53.5 & 54.3 & 55.6 & 56.9 & 75.9 & 80.0 & 81.2 & 82.5 & 83.3 & 50.3 & 51.0 & 51.2 & 52.3 & 52.7 & 50.8 & 51.8 & 52.4 & 53.3 & 53.7 \\
\rowcolor{gray!30}
Entropy & 50.9 & 52.1 & 52.6 & 53.1 & 53.4 & 76.3 & 79.7 & 80.7 & 81.9 & 82.7 & 49.6 & 50.2 & 50.5 & 50.9 & 51.3 & 51.7 & 51.8 & 52.1 & 52.2 & 52.3 \\
\rowcolor{skyblue!30}
Reference & 51.7 & 54.4 & 55.2 & 57.4 & 58.5 & 37.3 & 41.0 & 41.8 & 43.2 & 43.6 & 50.9 & 52.7 & 52.8 & 53.9 & 54.5 & 49.4 & 52.2 & 52.6 & 53.5 & 54.1 \\
\rowcolor{skyblue!30}
Zlib & 50.4 & 53.1 & 54.0 & 55.7 & 56.7 & 79.7 & 82.9 & 83.9 & 85.0 & 85.7 & 51.1 & 52.1 & 52.3 & 53.2 & 53.6 & 51.5 & 52.6 & 53.1 & 53.7 & 54.2 \\
\rowcolor{skyblue!30}
Neighborhood & 51.2 & 54.4 & 54.8 & 56.0 & 57.7 & 75.4 & 74.7 & 74.1 & 74.8 & 74.9 & 51.4 & 52.7 & 53.3 & 54.8 & 54.7 & 52.6 & 55.0 & 55.8 & 56.5 & 57.0 \\
\rowcolor{skyblue!30}
Min-K\% & 50.6 & 53.5 & 54.7 & 56.7 & 57.8 & 75.2 & 79.8 & 81.0 & 82.5 & 83.4 & 50.7 & 51.4 & 51.6 & 52.5 & 52.9 & 51.4 & 52.6 & 53.0 & 53.9 & 54.8 \\
\rowcolor{skyblue!30}
Min-K\%++ & 51.2 & 55.2 & 56.4 & 59.6 & 60.7 & 73.2 & 78.2 & 79.7 & 81.1 & 82.4 & 50.9 & 52.4 & 52.2 & 54.0 & 54.7 & 50.6 & 52.2 & 52.8 & 54.3 & 55.2 \\
\rowcolor{skyblue!30}
ReCaLL & 50.5 & 54.2 & 54.6 & 57.2 & 57.7 & 72.2 & 77.3 & 79.6 & 80.6 & 81.9 & 48.2 & 49.4 & 50.7 & 51.5 & 51.2 & 52.1 & 52.6 & 55.1 & 54.9 & 56.0 \\
\rowcolor{skyblue!30}
DC-PDD & 52.4 & 53.9 & 54.5 & 55.8 & 56.4 & 82.1 & 85.2 & 86.2 & 86.9 & 87.6 & 50.8 & 52.6 & 52.7 & 53.3 & 53.6 & 50.5 & 51.7 & 52.5 & 52.9 & 53.2 \\
\rowcolor{lightcoral!30}
DetectGPT & 51.2 & 54.4 & 54.8 & 56.0 & 57.7 & 75.4 & 74.7 & 74.1 & 74.8 & 74.9 & 51.4 & 52.7 & 53.3 & 54.8 & 54.7 & 52.6 & 55.0 & 55.8 & 56.5 & 57.0 \\
\rowcolor{lightcoral!30}
Fast-DetectGPT & 51.9 & 54.9 & 56.3 & 60.0 & 62.9 & 57.8 & 67.2 & 69.6 & 71.4 & 72.3 & 51.8 & 54.2 & 53.9 & 55.6 & 56.1 & 49.1 & 51.7 & 52.8 & 55.1 & 56.4 \\
\rowcolor{lightcoral!30}
Binoculars & 51.7 & 55.2 & 56.7 & 58.5 & 60.6 & 71.8 & 77.2 & 74.5 & 80.3 & 81.7 & 51.2 & 53.6 & 54.5 & 55.1 & 55.0 & 50.4 & 52.4 & 53.0 & 55.2 & 55.9 \\
\rowcolor{lightcoral!30}
DetectLLM & 51.6 & 54.0 & 55.4 & 58.3 & 61.4 & 56.4 & 66.8 & 69.6 & 71.3 & 71.8 & 52.2 & 53.7 & 53.5 & 55.5 & 55.8 & 49.0 & 51.3 & 52.5 & 54.8 & 55.7 \\
\rowcolor{lightcoral!30}
Lastde++ & 50.8 & 54.2 & 55.7 & 59.3 & 61.4 & 52.8 & 64.7 & 66.8 & 68.6 & 69.7 & 50.9 & 52.3 & 52.7 & 54.6 & 54.5 & 50.7 & 52.2 & 53.0 & 54.6 & 56.4 \\
\toprule
\multicolumn{1}{c}{\multirow{2.5}{*}{\textbf{Method}}} & \multicolumn{5}{c}{\textbf{ArXiv}} & \multicolumn{5}{c}{DM Mathematics} & \multicolumn{5}{c}{\textbf{HackerNews}} & \multicolumn{5}{c}{\textbf{Avg. (textual domains)}}  \\
\cmidrule(lr){2-6} \cmidrule(lr){7-11} \cmidrule(lr){12-16} \cmidrule(lr){17-21} 
& 160M & 1.4B & 2.8B & 6.9B & 12B & 160M & 1.4B & 2.8B & 6.9B & 12B & 160M & 1.4B & 2.8B & 6.9B & 12B & 160M & 1.4B & 2.8B & 6.9B & 12B \\
\midrule 
\rowcolor{gray!30}
Loss & 54.6 & 55.8 & 56.4 & 57.5 & 58.1 & 67.8 & 67.5 & 67.2 & 67.3 & 67.3 & 50.5 & 51.8 & 52.6 & 53.4 & 54.2 & 51.5 & 52.8 & 53.4 & 54.4 & 55.1 \\
\rowcolor{gray!30}
Rank & 52.0 & 52.4 & 52.6 & 54.6 & 55.0 & 60.6 & 60.4 & 60.7 & 60.6 & 60.6 & 51.7 & 52.7 & 52.5 & 53.6 & 54.5 & 51.2 & 52.4 & 52.6 & 53.9 & 54.5 \\
\rowcolor{gray!30}
LogRank & 54.4 & 55.6 & 56.1 & 57.5 & 57.9 & 66.3 & 66.4 & 66.2 & 66.4 & 66.2 & 50.4 & 51.7 & 52.4 & 53.7 & 54.3 & 51.4 & 52.7 & 53.3 & 54.5 & 55.1 \\
\rowcolor{gray!30}
Entropy & 54.8 & 55.4 & 55.1 & 55.7 & 55.7 & 68.4 & 67.6 & 67.4 & 67.2 & 67.1 & 50.5 & 52.2 & 52.0 & 52.1 & 52.1 & 51.5 & 52.3 & 52.5 & 52.8 & 53.0\\
\rowcolor{skyblue!30}
Reference & 51.2 & 53.3 & 54.1 & 55.8 & 56.8 & 45.0 & 44.8 & 44.4 & 44.4 & 44.3 & 50.0 & 52.1 & 53.8 & 55.2 & 56.6 & 50.6 & 52.9 & 53.7 & 55.2 & 56.1 \\
\rowcolor{skyblue!30}
Zlib & 54.2 & 55.3 & 55.7 & 56.7 & 57.2 & 64.6 & 64.7 & 64.6 & 64.6 & 64.6 & 51.2 & 51.9 & 52.4 & 52.9 & 53.4 & 51.7 & 53.0 & 53.5 & 54.4 & 55.0\\
\rowcolor{skyblue!30}
Neighborhood & 53.1 & 54.7 & 54.2 & 54.9 & 55.0 & 53.3 & 51.8 & 53.1 & 50.8 & 53.3 & 51.1 & 51.0 & 51.8 & 51.9 & 52.6 & 51.0 & 52.2 & 52.6 & 53.4 & 53.9 \\
\rowcolor{skyblue!30}
Min-K\% & 53.3 & 55.0 & 55.8 & 57.3 & 58.4 & 64.7 & 65.2 & 64.9 & 65.1 & 65.1 & 50.9 & 51.8 & 53.1 & 54.3 & 55.3 & 51.4 & 52.9 & 53.6 & 54.9 & 55.8 \\
\rowcolor{skyblue!30}
Min-K\%++ & 51.3 & 53.8 & 55.9 & 56.9 & 59.9 & 58.8 & 57.9 & 58.7 & 58.4 & 58.2 & 51.1 & 51.5 & 53.1 & 54.6 & 56.4 & 51.0 & 53.0 & 54.1 & 55.9 & 57.4 \\
\rowcolor{skyblue!30}
ReCaLL & 53.4 & 54.5 & 55.4 & 57.3 & 58.3 & 58.0 & 56.4 & 56.8 & 53.7 & 53.1 & 52.6 & 52.4 & 52.4 & 53.7 & 54.0 & 51.4 & 52.6 & 53.6 & 54.9 & 55.4 \\
\rowcolor{skyblue!30}
DC-PDD & 54.7 & 56.3 & 56.3 & 57.4 & 57.7 & 63.9 & 63.8 & 63.5 & 63.4 & 63.6 & 49.4 & 51.0 & 51.4 & 52.1 & 52.7 & 51.6 & 53.1 & 53.5 & 54.3 & 54.7 \\
\rowcolor{lightcoral!30}
DetectGPT & 53.1 & 54.7 & 54.2 & 54.9 & 55.0 & 53.3 & 51.8 & 53.1 & 50.8 & 53.3 & 51.1 & 51.0 & 51.8 & 51.9 & 52.6 & 51.0 & 52.2 & 52.6 & 53.4 & 53.9 \\
\rowcolor{lightcoral!30}
Fast-DetectGPT & 51.5 & 53.2 & 54.9 & 57.4 & 59.3 & 52.4 & 53.1 & 54.2 & 54.0 & 54.7 & 50.1 & 49.5 & 51.9 & 54.1 & 56.4 & 50.9 & 52.7 & 54.0 & 56.4 & 58.2 \\
\rowcolor{lightcoral!30}
Binoculars & 54.1 & 54.6 & 54.9 & 57.6 & 59.9 & 55.7 & 53.8 & 53.5 & 52.1 & 52.8 & 49.6 & 50.4 & 51.5 & 53.8 & 55.6 & 51.3 & 53.3 & 54.2 & 56.1 & 57.5\\
\rowcolor{lightcoral!30}
DetectLLM & 51.5 & 53.1 & 54.7 & 57.6 & 58.6 & 51.9 & 51.8 & 53.2 & 53.6 & 53.0 & 49.8 & 49.2 & 51.1 & 53.6 & 55.5 & 50.8 & 52.3 & 53.4 & 56.0 & 57.4 \\
\rowcolor{lightcoral!30}
Lastde++ & 50.8 & 52.9 & 54.7 & 56.1 & 58.1 & 52.3 & 51.6 & 52.1 & 52.2 & 52.3 & 50.0 & 51.1 & 52.0 & 54.0 & 56.0 & 50.6 & 52.5 & 53.6 & 55.7 & 57.3 \\
\toprule
\end{tabular}
\label{mia_detection_performance_result_mimir}
\end{table*}

\begin{table*}[b]
\caption{Full comparison of white-box machine text detection performances (AUROC) of MIA methods and machine text detectors on the RAID benchmark. Target models are GPT-2: GPT-2-XL, MPT: MPT-30B-Chat, LLaMA: LLaMA-2-70B-Chat. Gray, blue, and red areas indicate general baseline methods, MIA methods, and machine text detectors, respectively.}
\centering
\fontsize{8pt}{8pt}\selectfont
\setlength{\tabcolsep}{2.5pt} 
\renewcommand{\arraystretch}{1.3} 
\begin{tabular}{lcccccccccccc}
\toprule
 \multicolumn{1}{c}{\multirow{2.5}{*}{\textbf{Method}}} & \multicolumn{3}{c}{\textbf{Abstracts}} & \multicolumn{3}{c}{\textbf{Books}} & \multicolumn{3}{c}{\textbf{News}} & \multicolumn{3}{c}{\textbf{Poetry}} \\
\cmidrule(lr){2-4} \cmidrule(lr){5-7} \cmidrule(lr){8-10} \cmidrule(lr){11-13} 
 & GPT-2 & MPT  & LLaMA & GPT-2 & MPT  & LLaMA & GPT-2 & MPT  & LLaMA & GPT-2 & MPT  & LLaMA \\
\midrule
\rowcolor{gray!30}
Loss & 97.4 & 98.6 & 100.0 & 98.6 & 99.9 & 100.0 & 94.6 & 99.9 & 100.0 & 96.8 & 98.3 & 96.2 \\
\rowcolor{gray!30}
Rank & 98.2 & 53.9  & 93.8 & 99.4 & 91.9 & 96.5 & 95.5 & 70.3 & 86.1 & 98.0 & 85.6  & 88.8 \\
\rowcolor{gray!30}
LogRank & 98.5 & 96.4  & 100.0 & 99.4 & 99.9 & 100.0 & 96.5 & 99.7  & 100.0 & 97.6 & 98.3  & 95.6 \\
\rowcolor{gray!30}
Entropy & 61.6 & 55.0  & 99.9 & 60.7 & 97.4  & 99.9 & 38.9 & 92.3 & 99.9 & 89.3 & 90.3  & 95.4 \\
\rowcolor{skyblue!30}
Reference & 21.7 & 52.4 & 53.5 & 59.7 & 81.7 & 65.2 & 45.9 & 97.8 & 81.4 & 40.8 & 88.7 & 73.6 \\
\rowcolor{skyblue!30}
Zlib & 87.1 & 47.0 & 99.9 & 64.1 & 66.8 & 96.0 & 55.8 & 80.4 & 99.7 & 73.4 & 73.9 & 87.6 \\
\rowcolor{skyblue!30}
Neighborhood & 95.3 & 58.4 & 94.7 & 92.1 & 64.8  & 86.3 & 85.3 & 67.7  & 88.9 & 83.0 & 73.2  & 76.2 \\
\rowcolor{skyblue!30}
Min-K\% & 99.8 & 98.8  & 100.0 & 99.9 & 99.9 & 100.0 & 99.4 & 99.9  & 100.0 & 99.3 & 99.0  & 96.8 \\
\rowcolor{skyblue!30}
Min-K\%++ & 99.6 & 99.6  & 97.1 & 99.8 & 99.0  & 98.2 & 99.7 & 99.5  & 98.9 & 99.4 & 99.1 & 90.3 \\
\rowcolor{skyblue!30}
ReCaLL & 62.6 & 87.1  & 99.6 & 81.1 & 95.3  & 97.8 & 68.8 & 89.9  & 93.3 & 76.9 & 87.7  & 97.6 \\
\rowcolor{skyblue!30}
DC-PDD & 48.6 & 95.3  & 100.0 & 64.1 & 99.8  & 99.8 & 47.4 & 99.3  & 99.9 & 82.5 & 97.0 & 94.2 \\
\rowcolor{lightcoral!30}
DetectGPT & 95.3 & 58.4 & 94.7 & 92.1 & 64.8  & 86.3 & 85.3 & 67.7  & 88.9 & 83.0 & 73.2  & 76.2 \\
\rowcolor{lightcoral!30}
Fast-DetectGPT & 99.7 & 100.0  & 99.8 & 99.8 & 100.0 & 99.9 & 99.7 & 100.0  & 100.0 & 99.3 & 100.0  & 94.4 \\
\rowcolor{lightcoral!30}
Binoculars & 99.8 & 99.9  & 100.0 & 99.8 & 100.0  & 100.0 & 99.8 & 99.9  & 100.0 & 99.5 & 100.0 & 100.0 \\
\rowcolor{lightcoral!30}
DetectLLM & 99.7 & 100.0 & 100.0 & 99.8 & 100.0  & 99.8 & 99.7 & 100.0  & 100.0 & 99.4 & 99.9  & 93.2 \\
\rowcolor{lightcoral!30}
Lastde++ & 99.7 & 100.0  & 97.8 & 99.8 & 100.0 & 98.8 & 99.6 & 100.0  & 100.0 & 99.4 & 99.9  & 88.8 \\
\toprule
 \multicolumn{1}{c}{\multirow{2.5}{*}{\textbf{Method}}} & \multicolumn{3}{c}{\textbf{Recipes}} & \multicolumn{3}{c}{\textbf{Reddit}} & \multicolumn{3}{c}{\textbf{Reviews}} & \multicolumn{3}{c}{\textbf{Wikipedia}} \\
\cmidrule(lr){2-4} \cmidrule(lr){5-7} \cmidrule(lr){8-10} \cmidrule(lr){11-13} 
 & GPT-2 & MPT  & LLaMA & GPT-2 & MPT  & LLaMA & GPT-2 & MPT  & LLaMA & GPT-2 & MPT  & LLaMA \\
\midrule
\rowcolor{gray!30}
Loss &  70.4 & 100.0& 100.0 & 97.7 & 97.4 & 99.7 & 97.9 & 99.9  & 99.9 & 91.4 & 100.0 & 99.9 \\
\rowcolor{gray!30}
Rank &  82.3 & 48.8 & 17.5 & 98.3 & 73.9  & 64.6 & 98.8 & 82.8  & 95.3 & 97.0 & 73.4  & 81.3 \\
\rowcolor{gray!30}
LogRank & 71.5 & 99.9  & 100.0 & 98.3 & 96.2  & 99.7 & 98.8 & 99.9  & 99.9 & 95.5 & 99.9  & 99.8 \\
\rowcolor{gray!30}
Entropy & 40.2 & 98.4  & 100.0 & 72.3 & 78.3  & 99.7 & 59.0 & 98.6  & 99.9 & 37.6 & 94.5  & 99.1 \\
\rowcolor{skyblue!30}
Reference & 34.9 & 84.7 & 12.6 & 51.8 & 53.2 & 85.4 & 60.4 & 78.0  & 85.9 & 46.8 & 92.6 & 64.9 \\
\rowcolor{skyblue!30}
Zlib & 61.0 & 98.0  & 99.9 & 96.0 & 50.9  & 99.6 & 60.3 & 65.9  & 99.2 & 87.4 & 99.3  & 99.8 \\
\rowcolor{skyblue!30}
Neighborhood  & 82.5 & 58.2  & 89.2 & 97.5 & 57.0  & 92.1 & 90.6 & 70.8  & 97.2 & 93.7 & 82.4 & 88.6 \\
\rowcolor{skyblue!30}
Min-K\% & 89.4 & 100.0  & 100.0 & 99.3 & 96.6  & 99.7 & 99.7 & 99.9  & 100.0 & 99.4 & 100.0  & 99.9 \\
\rowcolor{skyblue!30}
Min-K\%++ & 98.9 & 98.8  & 99.2 & 99.3 & 96.9  & 91.7 & 99.4 & 99.7  & 99.5 & 99.6 & 99.6  & 99.7 \\
\rowcolor{skyblue!30}
ReCaLL & 60.2 & 97.1  & 100.0 & 40.5 & 87.8  & 98.6 & 67.4 & 89.4  & 95.8 & 37.0 & 71.1  & 97.9 \\
\rowcolor{skyblue!30}
DC-PDD & 41.3 & 99.9  & 100.0 & 62.9 & 97.7  & 99.9 & 62.2 & 99.8  & 99.9 & 38.3 & 98.3  & 98.9 \\
\rowcolor{lightcoral!30}
DetectGPT  & 82.5 & 58.2  & 89.2 & 97.5 & 57.0  & 92.1 & 90.6 & 70.8  & 97.2 & 93.7 & 82.4 & 88.6 \\
\rowcolor{lightcoral!30}
Fast-DetectGPT & 99.1 & 100.0  & 100.0 & 99.8 & 99.9  & 94.5 & 99.6 & 99.8  & 99.6 & 99.8 & 100.0  & 100.0 \\
\rowcolor{lightcoral!30}
Binoculars & 99.6 & 100.0  & 100.0 & 99.7 & 99.1  & 100.0 & 99.8 & 99.9  & 100.0 & 99.8 & 100.0  & 100.0 \\
\rowcolor{lightcoral!30}
DetectLLM & 99.0 & 100.0  & 99.9 & 99.8 & 99.9  & 96.7 & 99.7 & 99.8  & 99.7 & 99.8 & 100.0  & 99.9 \\
\rowcolor{lightcoral!30}
Lastde++ & 99.0 & 100.0  & 99.4 & 99.8 & 99.6  & 85.1 & 99.5 & 99.6  & 98.0 & 99.9 & 100.0  & 99.9 \\
\toprule
\end{tabular}
\label{white_box_llm_detection_performance_result}
\end{table*}

\begin{table*}[t]
\caption{Full comparison of black-box machine text detection performances (AUROC) of MIA methods and machine text detectors on the RAID benchmark. Target models are ChatGPT and GPT-4. Surrogate models are Llama: LLaMA-3-3.2B, Pythia: \textsc{Pythia}-160M. Gray, blue, and red areas indicate general baseline methods, MIA methods, and machine text detectors, respectively.}
\centering
\fontsize{7pt}{7pt}\selectfont
\setlength{\tabcolsep}{2pt} 
\renewcommand{\arraystretch}{1.3} 
\begin{tabular}{lcccccccccccccccc}
\toprule
 \multicolumn{1}{c}{\multirow{4}{*}{\textbf{Method}}} & \multicolumn{4}{c}{\textbf{Abstracts}} & \multicolumn{4}{c}{\textbf{Books}} & \multicolumn{4}{c}{\textbf{News}} & \multicolumn{4}{c}{\textbf{Poetry}} \\
\cmidrule(lr){2-5} \cmidrule(lr){6-9} \cmidrule(lr){10-13} \cmidrule(lr){14-17} 
 & \multicolumn{2}{c}{\textbf{ChatGPT}}& \multicolumn{2}{c}{\textbf{GPT-4}} & \multicolumn{2}{c}{\textbf{ChatGPT}}& \multicolumn{2}{c}{\textbf{GPT-4}} & \multicolumn{2}{c}{\textbf{ChatGPT}}& \multicolumn{2}{c}{\textbf{GPT-4}} & \multicolumn{2}{c}{\textbf{ChatGPT}}& \multicolumn{2}{c}{\textbf{GPT-4}} \\
 \cmidrule(lr){2-3} \cmidrule(lr){4-5} \cmidrule(lr){6-7} \cmidrule(lr){8-9} \cmidrule(lr){8-9} \cmidrule(lr){10-11} \cmidrule(lr){12-13} \cmidrule(lr){14-15} \cmidrule(lr){16-17}
 & Llama & Pythia & Llama & Pythia & Llama & Pythia & Llama & Pythia & Llama & Pythia & Llama & Pythia & Llama & Pythia & Llama & Pythia \\
\midrule
\rowcolor{gray!30}
Loss & 96.0 & 95.6 & 38.5 & 51.9 & 99.8 & 98.2 & 79.1 & 72.0 & 99.5 & 99.1 & 66.1 & 84.0 & 88.9 & 79.4 & 46.0 & 51.0\\
\rowcolor{gray!30}
Rank & 73.6 & 82.9 & 49.1 & 57.3 & 99.5 & 91.8 & 96.0 & 59.3 & 93.2 & 92.3 & 62.6 & 69.2 & 85.6 & 73.9 & 48.0 & 36.5 \\
\rowcolor{gray!30}
LogRank & 96.6 & 96.2 & 40.3 & 53.8 & 99.8 & 97.5 & 77.6 & 65.1 & 99.7 & 98.8 & 68.1 & 80.4 & 88.9 & 76.1 & 44.1 & 43.0\\
\rowcolor{gray!30}
Entropy & 74.5 & 68.3 & 19.8 & 39.2 & 99.4 & 86.2 & 83.7 & 62.3 & 99.1 & 87.4 & 68.0 & 69.3 & 87.4 & 62.3 & 44.7 & 37.3 \\
\rowcolor{skyblue!30}
Reference & 37.6 & 50.3 & 29.0 & 38.9 & 58.4 & 89.2 & 50.6 & 70.4 & 50.0 & 54.8 & 33.8 & 42.1 & 79.5 & 77.2 & 71.4 & 50.1 \\
\rowcolor{skyblue!30}
Zlib & 53.3 & 39.7 & 12.4 & 14.8 & 72.5 & 59.8 & 64.1 & 62.2 & 89.6 & 76.9 & 70.5 & 72.7 & 68.2 & 56.0 & 60.2 & 63.4 \\
\rowcolor{skyblue!30}
Neighborhood  & 61.0 & 77.1 & 29.4 & 45.9 & 78.9 & 87.4 & 57.4 & 58.8 & 81.6 & 88.3 & 53.9 & 61.4 & 74.5 & 78.0 & 66.7 & 61.8\\
\rowcolor{skyblue!30}
Min-K\% & 98.3 & 97.5 & 56.1 & 62.5 & 99.8 & 96.4 & 74.4 & 58.6 & 99.6 & 98.5 & 63.5 & 76.3 & 89.8 & 77.0 & 42.3 & 37.2 \\
\rowcolor{skyblue!30}
Min-K\%++ & 99.1 & 99.3 & 71.9 & 68.7 & 90.3 & 98.9 & 36.7 & 70.4 & 73.6 & 99.7 & 38.0 & 86.3 & 65.5 & 92.4 & 44.4 & 69.0 \\
\rowcolor{skyblue!30}
ReCaLL & 99.3 & 98.0 & 81.1 & 97.4 & 96.3 & 87.6 & 65.6 & 83.2 & 87.4 & 66.5 & 47.0 & 55.7 & 89.0 & 73.8 & 52.3 & 54.5 \\
\rowcolor{skyblue!30}
DC-PDD & 88.7 & 85.9 & 38.5 & 62.5 & 99.0 & 96.1 & 77.2 & 86.0 & 93.6 & 85.1 & 47.6 & 72.0 & 87.2 & 80.9 & 51.0 & 64.9 \\
\rowcolor{lightcoral!30}
DetectGPT  & 61.0 & 77.1 & 29.4 & 45.9 & 78.9 & 87.4 & 57.4 & 58.8 & 81.6 & 88.3 & 53.9 & 61.4 & 74.5 & 78.0 & 66.7 & 61.8\\
\rowcolor{lightcoral!30}
Fast-DetectGPT & 99.3 & 98.5 & 77.0 & 58.7 & 83.9 & 96.1 & 38.0 & 69.9 & 72.2 & 99.4 & 46.5 & 88.0 & 70.9 & 89.7 & 53.2 &  79.0 \\
\rowcolor{lightcoral!30}
Binoculars & 100.0 & 98.1 & 96.5 & 94.5 & 99.4 & 97.1 & 77.0 & 70.5 & 98.8 & 78.6 & 83.0 & 51.2 & 98.9 & 85.9 & 57.4 & 73.0 \\
\rowcolor{lightcoral!30}
DetectLLM & 98.9 & 98.0 & 77.1 & 57.7 & 85.6 & 94.9 & 42.7 & 66.7 & 77.8 & 99.1 & 51.0 & 85.2 & 75.7 & 88.6 & 56.0 & 74.8 \\
\rowcolor{lightcoral!30}
Lastde++ & 98.6 & 96.9 & 75.8 & 58.3 & 83.6 & 91.8 & 40.2 & 60.6 & 69.1 & 98.6 & 45.2 & 84.2 & 71.9 & 84.7 & 53.0 & 67.0 \\
\toprule
 \multicolumn{1}{c}{\multirow{4}{*}{\textbf{Method}}} & \multicolumn{4}{c}{\textbf{Recipes}} & \multicolumn{4}{c}{\textbf{Reddit}} & \multicolumn{4}{c}{\textbf{Reviews}} & \multicolumn{4}{c}{\textbf{Wikipedia}} \\
\cmidrule(lr){2-5} \cmidrule(lr){6-9} \cmidrule(lr){10-13} \cmidrule(lr){14-17} 
 & \multicolumn{2}{c}{\textbf{ChatGPT}}& \multicolumn{2}{c}{\textbf{GPT-4}} & \multicolumn{2}{c}{\textbf{ChatGPT}}& \multicolumn{2}{c}{\textbf{GPT-4}} & \multicolumn{2}{c}{\textbf{ChatGPT}}& \multicolumn{2}{c}{\textbf{GPT-4}} & \multicolumn{2}{c}{\textbf{ChatGPT}}& \multicolumn{2}{c}{\textbf{GPT-4}} \\
 \cmidrule(lr){2-3} \cmidrule(lr){4-5} \cmidrule(lr){6-7} \cmidrule(lr){8-9} \cmidrule(lr){8-9} \cmidrule(lr){10-11} \cmidrule(lr){12-13} \cmidrule(lr){14-15} \cmidrule(lr){16-17}
 & Llama & Pythia & Llama & Pythia & Llama & Pythia & Llama & Pythia & Llama & Pythia & Llama & Pythia & Llama & Pythia & Llama & Pythia \\
\midrule
\rowcolor{gray!30}
Loss &  99.7 & 99.0 & 93.4 & 94.8 & 98.8 & 96.6 & 91.6 & 87.0 & 99.9  & 99.5 & 89.0 & 80.9 & 99.0 & 97.1 & 73.8 & 82.4  \\
\rowcolor{gray!30}
Rank &  56.9 & 89.7 & 48.0 & 90.5 & 87.7 & 86.2 & 79.0 & 75.7 & 99.1 & 94.7 & 75.7 & 63.1 & 94.8 & 93.3 & 77.2 & 75.4  \\
\rowcolor{gray!30}
LogRank & 99.5 & 98.9 & 92.9 & 94.6 & 98.5 & 95.6 & 90.4 & 84.3 & 99.9 & 99.2 & 86.2 & 72.2 & 99.2 & 97.4 & 76.0 & 82.3 \\
\rowcolor{gray!30}
Entropy & 99.7 & 92.6 & 93.5 & 83.1 & 96.5 & 88.5 & 90.8 & 82.1 & 99.9 & 93.4 & 91.2 & 66.0 & 98.7 & 83.4 & 73.6 & 72.9  \\
\rowcolor{skyblue!30}
Reference & 55.6 & 36.5 & 36.5 & 29.4 & 60.5 & 83.3 & 52.4 & 74.6 & 76.8 & 94.8 & 58.8 & 82.4 & 52.4 & 50.0 & 41.3 & 42.9 \\
\rowcolor{skyblue!30}
Zlib &  96.9 & 93.4 & 90.9 & 90.8 & 78.0 & 64.9 & 94.0 & 92.8 & 91.2 & 75.8 & 68.5 & 65.3 & 99.9 & 99.9 & 98.0 & 98.6 \\
\rowcolor{skyblue!30}
Neighborhood  & 66.3 & 87.5 & 58.6 & 78.8 & 76.4 & 86.6 & 74.6 & 72.1 & 94.3 & 87.6 & 72.7 & 55.3 & 92.9 & 91.7 & 71.7 & 68.0 \\
\rowcolor{skyblue!30}
Min-K\% &  99.5 & 98.7 & 92.8 & 93.9 & 98.3 & 93.9 & 87.5 & 79.6 & 99.9 & 98.1 & 79.5 & 65.4 & 99.5 & 97.9 & 78.9 & 81.2 \\
\rowcolor{skyblue!30}
Min-K\%++ &  95.1 & 99.7 & 76.6 & 94.5 & 89.0 & 96.8 & 64.8 & 78.6 & 91.5 & 99.2 & 46.0 & 81.6 & 84.5 & 99.3 & 52.3 & 80.8 \\
\rowcolor{skyblue!30}
ReCaLL & 95.9 & 88.0 & 69.2 & 74.2 & 97.0 & 77.3 & 58.2 & 43.2 & 87.9 & 65.0 & 58.4 & 67.7 & 80.3 & 28.0 & 34.6 & 22.9  \\
\rowcolor{skyblue!30}
DC-PDD & 98.2 & 94.4 & 88.1 & 84.7 & 98.6 & 96.0 & 87.3 & 86.2 & 99.8 & 97.0 & 86.7 & 90.8 & 88.3 & 71.9 & 45.2 & 55.2 \\
\rowcolor{lightcoral!30}
DetectGPT  & 66.3 & 87.5 & 58.6 & 78.8 & 76.4 & 86.6 & 74.6 & 72.1 & 94.3 & 87.6 & 72.7 & 55.3 & 92.9 & 91.7 & 71.7 & 68.0 \\
\rowcolor{lightcoral!30}
Fast-DetectGPT & 84.1 & 99.8 & 67.2 & 98.8 & 85.8 & 94.1 & 66.3 & 85.0 & 81.2 & 99.8 & 40.4 & 86.6 & 79.8 & 99.9 & 54.9 & 93.0 \\
\rowcolor{lightcoral!30}
Binoculars & 99.9 & 98.2 & 97.8 & 94.7 & 99.6 & 91.3 & 94.6 & 76.1  & 99.5 & 96.5 & 84.0 & 83.8 & 98.8 & 97.7 & 85.1 & 77.8 \\
\rowcolor{lightcoral!30}
DetectLLM & 82.2 & 99.8 & 66.6 & 99.1 & 85.9 & 92.6 & 67.4 & 83.6  & 84.6 & 99.7 & 42.2 & 84.8 & 82.1 & 99.9 & 56.8 & 92.6 \\
\rowcolor{lightcoral!30}
Lastde++ & 83.9 & 99.4 & 67.8 & 97.9 & 83.9 & 90.4 & 66.6 & 83.9 & 84.2 & 99.3 & 40.9 & 80.4 & 81.6 & 99.8 & 57.4 & 92.0 \\
\toprule
\end{tabular}
\label{black_box_llm_detection_performance_result}
\end{table*}

\end{document}